\documentclass{article}

\usepackage[final]{neurips_2019}

\usepackage[utf8]{inputenc}
\usepackage[T1]{fontenc}
\usepackage{amsmath}
\usepackage{amsfonts}
\usepackage{amssymb}
\usepackage{hyperref}
\usepackage{url}
\usepackage{booktabs}
\usepackage{nicefrac}
\usepackage{microtype}
\usepackage{algorithm}
\usepackage{algorithmic}
\usepackage{wrapfig}

\usepackage{multicol}

\usepackage{microtype}
\usepackage{graphicx}
\usepackage{subfigure}
\usepackage{booktabs}
\usepackage{natbib}

\newcommand*\sq{\mathbin{\vcenter{\hbox{\rule{.9ex}{.9ex}}}}}

\usepackage[T1]{fontenc}

\usepackage{algorithm}
\usepackage{algorithmic}

\usepackage{color, colortbl}
\usepackage[table,x11names]{xcolor}
\definecolor{light-gray}{gray}{0.95}
\usepackage{tikz}

\usepackage{hyperref}

\DeclareMathOperator*{\argmax}{arg\,max}
\DeclareMathOperator*{\argmin}{arg\,min}

\let\emptyset\varnothing

\title{Meta-Learning surrogate models for sequential decision making}

\author{
 Alexandre Galashov\thanks{Joint first authorship, $\dag$ Joint senior authorship}, Jonathan Schwarz\footnotemark[1] \\
 \textbf{Hyunjik Kim}, \textbf{Marta Garnelo}, \textbf{David Saxton}, \textbf{Pushmeet Kohli},\\ 
 \textbf{S.M. Ali Eslami}\footnotemark[2], \textbf{Yee Whye Teh}\footnotemark[2]\\
 \texttt{\{agalashov, schwarzjn\}@google.com} \\
 DeepMind\\
 London, United Kingdom\\
}

\begin{document}

\maketitle

\begin{abstract}
 We introduce a unified probabilistic framework for solving sequential decision making problems ranging from Bayesian optimisation to contextual bandits and reinforcement learning. This is accomplished by a probabilistic model-based approach that explains observed data while capturing predictive uncertainty during the decision making process. Crucially, this probabilistic model is chosen to be a Meta-Learning system that allows learning from a distribution of related problems, allowing data efficient adaptation to a target task. 
 As a suitable instantiation of this framework, we explore the use of Neural processes due to statistical and computational desiderata. We apply our framework to a broad range of problem domains, such as control problems, recommender systems and adversarial attacks on RL agents, demonstrating an efficient and general black-box learning approach.
\end{abstract}

\section{Introduction}
\label{sec:intro}

Sequential decision making encompasses a large range of problems with many decades of research targeted at problems such as Bayesian optimisation \citep[e.g.][]{mockus1978application, schonlau1998global}, contextual bandits \citep[e.g.][]{cesa2006prediction} and reinforcement learning. While recent years have brought great advances, allowing the successful application to increasingly complex decision making problems, most modern algorithms still require multiple magnitudes more experience than humans to solve even relatively simple problems.

For example, consider the task of designing a motor controller for an array of robot arms in a large factory. The robots vary in age, size and proportions. The objective of the controller is to send motor commands to the robot arms in such a way that allows each arm to achieve its designated task. The majority of current methods may tackle the control of each arm as a separate problem, despite similarity between arms and their respectively assigned tasks. Instead, we argue that availability of data on related problems ought to be harnessed and discuss how learning \textit{data-driven priors} can allow fast customisation of a general controller to additional robot arms in a fraction of the time.

A second issue that arises in the design of such controller is how to deal with uncertainty, e.g.\ uncertainty about the proportions of each robot, the physics of their motor movements and the state of the environment. While there exist several probabilistic methods for decision making that model predictive uncertainty (e.g. Gaussian Processes) and can be combined with techniques to balance exploration with exploitation, such methods may require significant domain-knowledge for appropriate calibration. In addition, most popular approaches cannot transfer knowledge between learning processes unless problem-specific modifications are being made, limiting the generality of the resulting framework.
 
In this paper, we argue that the availability of data from related tasks allow the application of algorithms specifically designed to learn from such distributions of related problems, shifting the paradigm from \textit{hand-designed} to \textit{data-driven} priors. Provided appropriate care is taken to allow reasoning about model-uncertainty, such techniques can in turn be combined with efficient exploration techniques, making them highly competitive and natural choices for decision making problems. Specifically, we argue that an efficient and fully automated framework for sequential decision making should have the following properties: (i) \textbf{Statistical efficiency:} Accurate predictions of function values based on small numbers of evaluations and
(ii) \textbf{Calibrated uncertainties:} To balance exploration and exploitation. We argue that such a set of methods can be obtained by learning from distribution of related tasks.

To this end, we introduce a probabilistic framework based on the ideas surrounding the active field of Meta-Learning \cite[e.g.][]{schmidhuber1987evolutionary, finn2017model}, showing how modern Meta-Learning techniques can be employed to decision making problems with minimal overhead. 

\section{Meta-Learning for sequential decision making}

We now discuss several instantiations of the general decision making problem, showing how Meta-Learning techniques can be applied in each instance. In all cases, we will make choices under uncertainty to optimise some notion of utility. Throughout this section we will use $\mathcal{M}_{\theta}$ with parameters $\theta$ to denote a model of the problem at hand. Furthermore, as the central paradigm of this paper, we assume the existence of some task distribution $p(\mathcal{T})$, related to a held-out target problem $\mathcal{T}^*$ of interest. Note that, as we simply assume $\mathcal{M}$ to be a general regression algorithm, the vast majority of Meta-Learning techniques developed for supervised learning may directly be used as instances of $\mathcal{M}$ and are thus applicable in the problems below. However, we specifically advocate for probabilistic methods due to the exploration problem. We thus draw some function from $\mathcal{M}$, denoted $\hat{g} \sim \mathcal{M}$. In a Bayesian setting for instance, this corresponds to drawing $\widetilde{\theta} \sim p(\theta|\mathcal{D})$ from the posterior over parameters and making predictions with $\mathcal{M}$ parameterised by $\widetilde{\theta}$.

\subsection{Bayesian Optimisation}
\label{sec:bayesopt}

\begin{wrapfigure}{L}{0.55\textwidth}
\vspace{-20pt}
\begin{minipage}{0.55\textwidth}
    \begin{algorithm}[H]
        \caption{Bayesian Optimisation}
        \label{alg:bo}
        \begin{algorithmic}
          \STATE {\bfseries Input:}
          \STATE $f^*$ - Target function of interest (= $\mathcal{T}^*$).
          \STATE $\mathcal{D}_{0} = \{(x_{0}, y_{0})\}$ - Observed evaluations of $f^*$.
          \STATE $N$ - Maximum number of function iterations.
          \STATE $\mathcal{M}_{\theta}$ - Model pre-trained on evaluations of similar functions $f_1, \dots f_n \sim p(\mathcal{T})$.
          \STATE
          \FOR{n=1, \ldots, N}
          \STATE \texttt{// Model-adaptation}
          \STATE Optimise $\theta$ to improve $\mathcal{M}$'s prediction on $\mathcal{D}_{n-1}$.
          \STATE
          \STATE \textit{Thompson sampling}: Draw $\hat{g}_{n} \sim \mathcal{M}$, find
          \STATE
          \STATE 
          $\qquad\qquad x_{n} = \argmin_{x \in \mathcal{X}} \mathbb{E} \big[\hat{g}(y|x)\big]$
          \STATE 
          \STATE Evaluate target function and save result.
          \STATE $\mathcal{D}_n \leftarrow \mathcal{D}_{n-1} \cup \{(x_{n}, f^*(x_{n}))\}$
          \ENDFOR
        \end{algorithmic}
    \end{algorithm}
\end{minipage}
\vspace{-20pt}
\end{wrapfigure}

We first consider the problem of optimising black-box functions without gradient information. A popular approach is Bayesian Optimisation (BO) \citep[e.g.][]{shahriari2016taking}, where we are to find the minimiser $x^* = \argmin_{x\in \mathcal{X}} f^*(x)$ of some function $f^*$ on $\mathcal{X}$ without requiring access to its derivatives. The BO approach consists of fitting a probabilistic surrogate model to approximate $f^*$ on a small set of evaluations $\mathcal{D} = \{(x_i, y_i)\}$ observed thus far. Examples of a surrogate are Gaussian Processes or Tree-structured Parzen (density) estimators. The decisions involved in the process is the choice of some $x'$ at which we choose to next evaluate the function $f^*$. This evaluation is typically assumed to be costly, e.g. when the optimisation of an algorithm is involved \citep{snoek2012practical}.

The key to the application of Meta-Learning techniques to BO is the task-distribution $p(\mathcal{T})$, which we assume to cover similar functions (e.g. in terms of the function domains, smoothness assumptions etc.). Thus, in order to transfer knowledge, we will learn a model $\mathcal{M}$ of $f^*$ by first pre-training on some available draws $f_1, \dots, f_n \sim p(\mathcal{T})$ (thus estimating properties of $f^*$) \textit{prior} to adapting the method to a typically much smaller set of evaluations of $f^*$.

In addition to providing a good approximation from limited data, we require $\mathcal{M}$ to provide uncertainty estimates, which is helpful in addressing the inherent exploration/exploitation trade-off in decision making. Thus we formulate an acquisition function $\alpha: \mathcal{X} \rightarrow \mathbb{R}$ to guide decision making, designed such that we consider $x = \argmax_{x'} \alpha(x')$ at the next point for evaluation. Model uncertainty is typically incorporated into $\alpha$, as is done in popular choices such as expected improvement \citep{mockus1978application} or the UCB algorithm \citep{srinivas2009gaussian}. Throughout this paper, we will use the Thompson sampling \citep{thompson1933likelihood} criterion. That is, $x$ is chosen for evaluation with probability

\begin{equation}
\mathbb{E}_{\hat{g}\sim\mathcal{M}}\Bigg[ \mathbb{I}\bigg[ \mathbb{E}[\hat{g}(y|x)] = \min_{x'} \mathbb{E}[\hat{g}(y|x')] \bigg]\Bigg]
\label{eq:full_thomposon_sampling}
\end{equation}

which we approximate by drawing a single $\hat{g} \sim \mathcal{M}$ and choosing its minimum as the next $x'$ for evaluation. Importantly, after each evaluation we adjust the model $\mathcal{M}$ to provide a good fit to all function evaluations obtained thus far (e.g. by backpropagation for optimisation-based Meta-Learning). This procedure is shown in Algorithm~\ref{alg:bo}.

\subsection{Contextual Multi-armed Bandits}
\label{sec:bandits}

Closely related to Bayesian Optimisation, the decision problem known as a contextual multi-armed bandit is formulated as follows. At each trial $t$:

\begin{enumerate}
\item Some context $s_t$ is revealed. This could be features describing a user of an online content provider. Crucially, we assume $s_t$ to be independent of past trials.

\item Next, we are to choose one of $k$ arms $a^1, \dots, a^k \in \mathcal{A}$ and receive reward $r_t \sim p_{a_t}$. The current context $s_t$, past actions and rewards $(s_{\tau}, r_{\tau})_{\tau=1}^{t-1}$ are available to guide this choice. As $\mu_k = \mathbb{E}_{r\sim p_{a_k}}[r]$ is unknown, we face the same exploration/exploitation trade-off.

\item Model-adaptation: The arm-selection strategy is updated given access to the newly acquired $(s_{t}, r_{t})$. Importantly, no reward is provided for any of the arms $a \neq a_t$.
\end{enumerate}

Given the generality of $\mathcal{M}$, the conceptual difference between the BO case in our framework is relatively minor, as we merely replace all occurrences of $g(y|x)$ in Algorithm \ref{alg:bo} by $g(r_t|x_t, s_t, a_t)$, i.e.
we evaluate $g$ separately for each arm. Assuming for instance that $g$ is a neural network, $x_t$ and $s_t$ can be concatenated. Thereafter, we choose the next arm to evaluate and proceed as before.

\subsection{Model-based Reinforcement Learning}
\label{sec:mbrl_results}

Allowing for dependence between subsequently provided contexts (referred to as states $s \in \mathcal{S}$ in the reinforcement learning (RL) literature) we arrive at RL \citep{sutton2018reinforcement}. An RL problem is defined by (possibly stochastic) functions $f_t: \mathcal{S}\times\mathcal{A} \rightarrow \mathcal{S}$ (defining the transitions between states given an agent's actions) and the reward function $f_r: \mathcal{S}\times\mathcal{A} \rightarrow \mathbb{R}$. These functions are together referred to as an environment. We obtain the necessary distribution over functions for pre-training by varying the properties of the environment, as before writing $p(\mathcal{T})$ to denote the distribution over functions for each task $\mathcal{T}_w$, i.e. $\mathcal{T}_w = (f^w_t, f^w_r) \sim p(\mathcal{T})$. The objective of the RL algorithm for a fixed task $w$ is $\max \mathbb{E}_{\pi}[\sum_{t \geq 0} \gamma^t r^{w}_t]$, for task-specific rewards $r^{w}$ obtained by acting on $\mathcal{T}_w$. $\pi: \mathcal{S} \rightarrow \mathcal{A}$ is a policy, i.e. the agent's decision making process. We also introduce $\gamma \in [0,1]$, a discounting factor and $t$, indicating a time index in the current episode.

In this paper, we will focus our attention to a particular set of techniques referred to as model-based algorithms. Model-based RL methods assume the existence of some approximation $\hat{f}_t, \hat{f}_r$ to the dynamics of the problem at hand (typically learned online). Examples of this technique are \citep[e.g.][]{peng1993efficient, browne2012survey}. Analogous to the BO and contextual bandit case, we tackle this problem by first Meta-Learning an environment model using some exploratory policy $\pi_\phi$ (e.g. a random walk or curiosity-driven algorithm) on samples of the task distribution $p(\mathcal{T})$. This gives us an environment model capable of quickly adapting to the dynamics of new problem instances. The focus on model-based RL techniques in conjunction with Meta-Learning is natural, in that data-efficiency is a main motivation in both cases.

Having pre-trained $\mathcal{M}$, we use the model in conjunction with any RL algorithm to learn a policy $\pi_{\psi}$ for the target task $\mathcal{T}^* \sim p(\mathcal{T})$. This can be done by autoregressively sampling rollouts from $\mathcal{M}$ (i.e. by acting according to $\pi_{\psi}$ and sampling transitions using $\mathcal{M}'s$ approximation to $\hat{f}^*_t, \hat{f}^{*}_{r}$). These rollouts are then used to update $\psi$ using any RL algorithm of choice. Optionally, we may also update $\psi$ using the real environment rollouts. We provide further algorithmic details in the Appendix.

Note that computational complexity is of particular importance in this problem: As we allow for additional episodes on the real environment, the number of transitions that could be added to a context set grows quickly ($\mathcal{O}(mk)$ for $m$ episodes of $k$ steps). In complex environments, this may quickly become prohibitive, e.g. for GP environment models \citep{deisenroth2011pilco}.

\section{Neural Processes as Meta-Learning models}
\label{sec:nps}

The framework for sequential decision making presented thus far is general and mostly model-agnostic, with little conceptual difference between a broad range of decision making problems. As a particular instance for model $\mathcal{M}$ in this paper, we use the recently introduced Neural processes (NPs) \citep{garnelo2018neural} but emphasise that other methods are straight-forwardly applicable. 

NPs are a family of neural models for few-shot learning, that given a number of realisations from some unknown stochastic process $f:\mathcal{X} \to \mathcal{Y}$, are trained to predict the values of $f$ at some new, unobserved locations. In contrast to standard approaches to supervised learning, NPs model a distribution over functions that agree with the observations provided so far (similar to e.g. GPs \citep{rasmussen2003gaussian}). NPs require a dataset of evaluations of similar functions $f_1, \dotsm f_n$ over the same spaces $\mathcal{X}$ and $\mathcal{Y}$. However, note that we do not assume each function to be evaluated at the same $x \in \mathcal{X}$. 

Importantly, Neural Processes naturally fit the criteria for decision making outlined in Section \ref{sec:intro}: (i) \textbf{Statistical efficiency} and (ii) \textbf{Calibrated uncertainties}. In addition, further desirable properties are: (i) The complexity of a Neural Process is $\mathcal{O}(n+m)$, allowing its application over long decision making processes (such as in the model-based RL case). (ii) Due to its non-parametric nature, no gradient steps are taken at test time, reducing the burden of the choice otherwise crucial hyperparameters during evaluation.\footnote{For popular optimised-based Meta-Learning techniques, e.g. \citep{finn2017model}, learning rates and update frequency must be specified, choices that have strong impact on the performance on $\mathcal{T}^*$.}. This also reduces computational cost.

Note that all code used to train neural processes for the purpose of this paper has been made available online by the authors \url{https://github.com/deepmind/neural-processes}.

\section{Experiments}
\label{sec:setup}

We now proceed to demonstrate our framework in conjunction with a Neural Process on a range of challenging problem. We strongly encourage the interested reader to consult the Appendix for the majority of experimental details due to space constraints.

\subsection{Model-based RL}
\label{sec:mbrl}

As a first example of using Meta-Learned surrogate models, we apply our method to the Cart-pole swing-up experiment proposed in \citep{saemundsson2018meta}. We obtain a distribution over tasks $p(\mathcal{T})$ (i.e. state transition and reward functions) by uniformly sampling the pole mass $p_m \sim \mathcal{U} [0.01, 1.0]$ and cart mass $c_m \sim \mathcal{U} [0.1, 3.0]$ for each episode, making pre-training of Meta-Learning methods possible. Note that this task distribution is noticeably broader than the original specification in \citep{saemundsson2018meta}, a change we introduced to increase complexity of an otherwise simple problem domain.

We compare the proposed method to a range of model-based techniques, namely MAML \footnote{We apply MAML on the model as opposed to (more commonly) the policy parameters.} \citep{finn2017model} and a Multi-task learning method, both also pre-trained on $p(\mathcal{T})$. In addition, we show results for competitive model-free algorithms. As the RL algorithm of choice for all model-based methods, we use on-policy SVG(1) \citep{heess2015learning} without replay.

\begin{figure*}[t!]
    \centering
    \subfigure[Example learning curves.\newline$\sq$ Model-based methods\newline$\bullet$ Model-free methods]{\label{fig:model_based_rl_results}\includegraphics[width=0.3\linewidth]{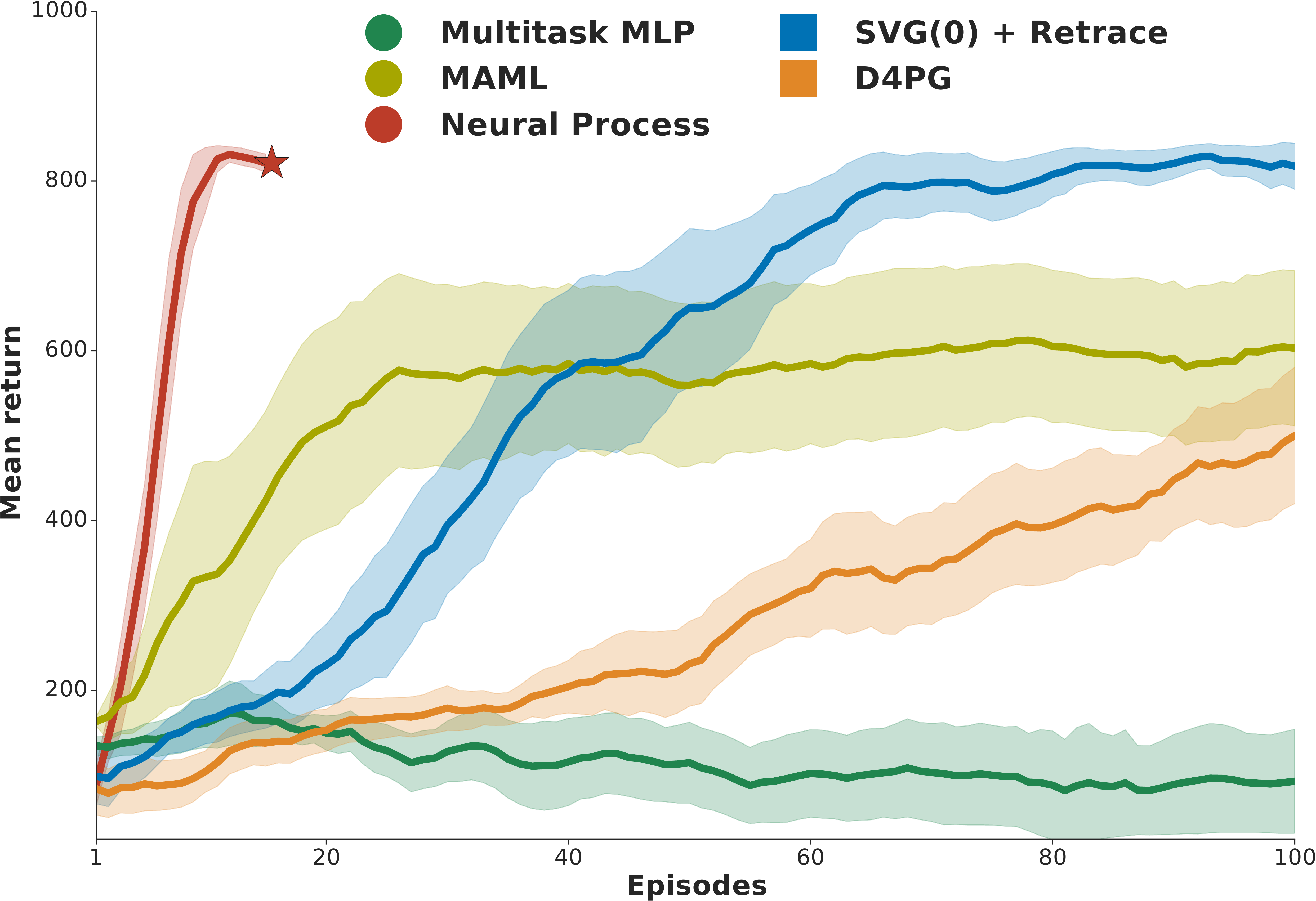}}
    \subfigure[Environment and model rollouts using a random initial state and $\pi_{\psi}$ .]{\label{fig:rollouts}\includegraphics[width=0.35\linewidth]{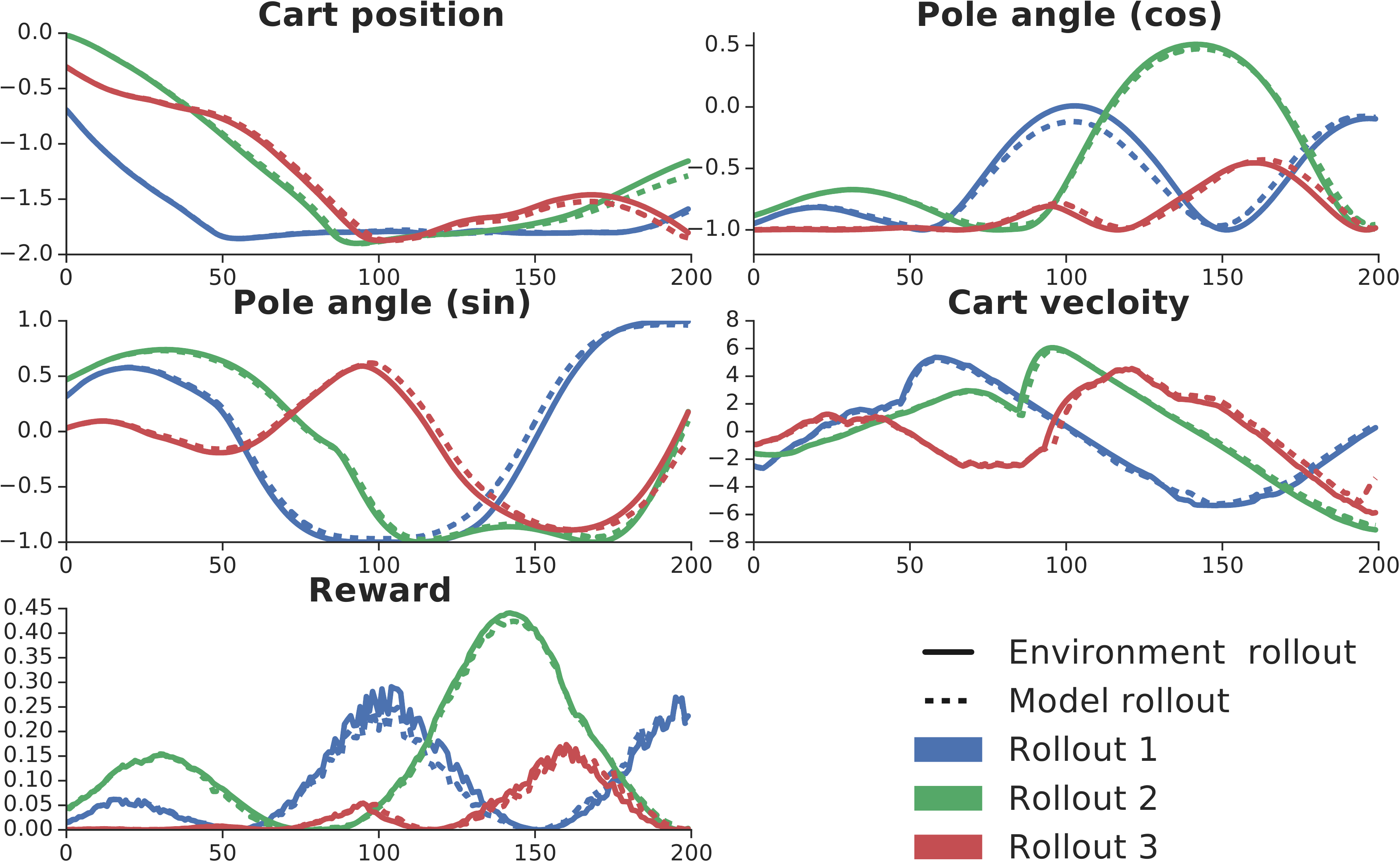}}
    \subfigure[Performance for different task parameters]{\label{fig:surface}\includegraphics[width=0.3\linewidth]{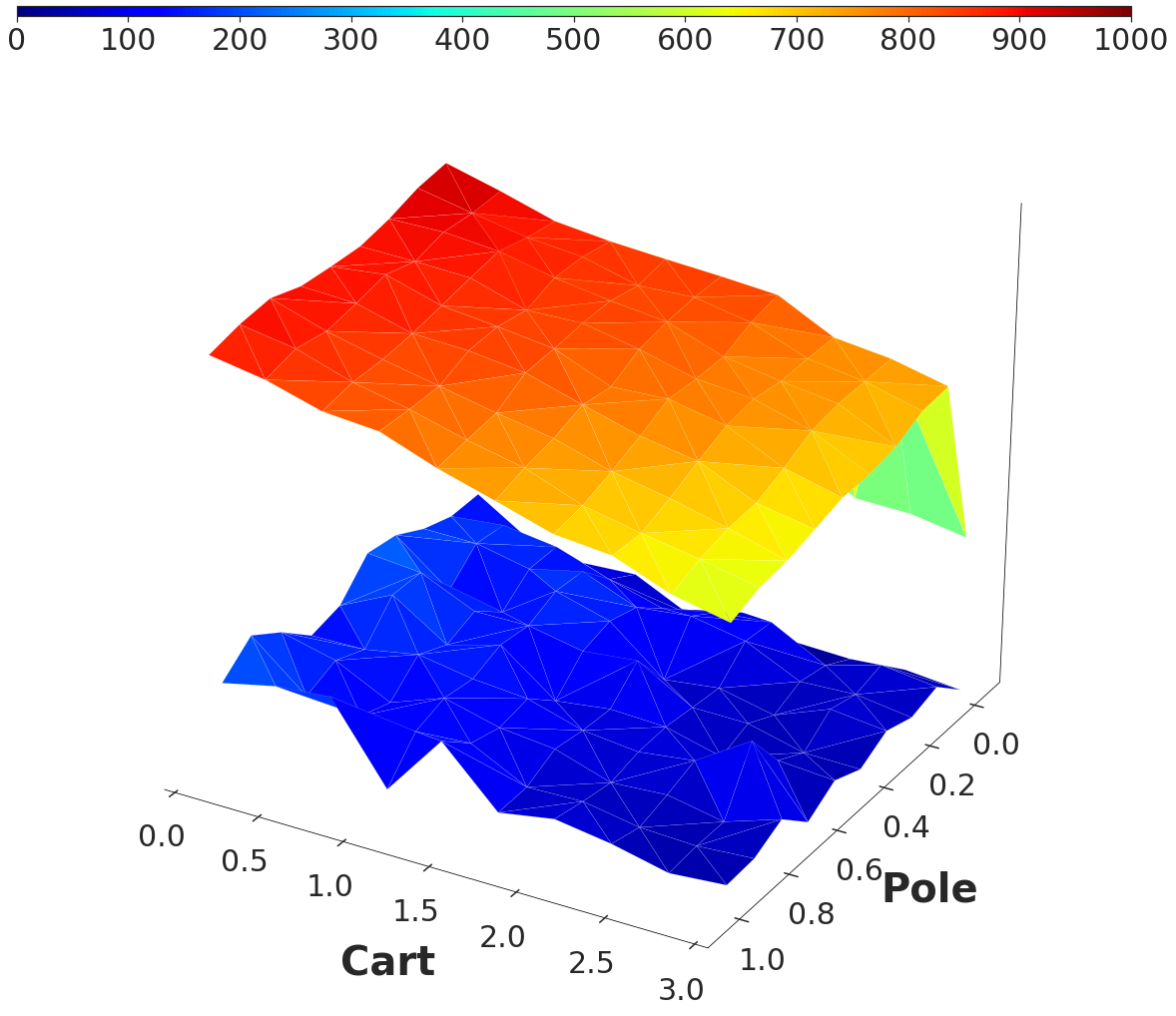}}
    \caption{\textbf{Left}: Learning curves (showing mean and half the standard deviation over 10 random repetitions.). \textbf{Middle:} Example rollouts for default task parameters $p_m=0.1, c_m=1.0$. \textbf{Right}: Mean episode reward at convergence for varying cart and pole masses.}
\end{figure*}

Results are shown in Figure \ref{fig:model_based_rl_results}. We observe strong performance, showing a model-based RL algorithm with a NP model can successfully learn a task in about 10-15 episodes.\footnote{An example video for a particular run can be found here: \url{https://tinyurl.com/y6p96t8m}}. Testing our method on the full support of the task distribution, we show the mean episode reward in Figure \ref{fig:surface} (comparing to a random policy in blue). We observe that the same method generalises for all considered tasks. As expected, the reward decreases slightly for particularly heavy carts. We also provide a comparison of NP rollouts comparing to the real environment rollouts in Figure ~\ref{fig:rollouts}.

\subsection{Recommender Systems}
\label{sec:method}

Considering next the contextual multi-armed bandit problem discussed in Section \ref{sec:bandits}, we apply our approach to recommender systems. As we aim to learn more about a user's preferences, we can think about certain recommendations as more exploratory in case they happen to be dissimilar to previously rated items.

The application of our framework to this problem is natural: We can think of each user $u$ as a function from items $\mathcal{I}$ to ratings $\mathcal{R}$, i.e. $f_u: \mathcal{I} \rightarrow \mathcal{R}$, where each user $f_u$ is possibly evaluated on a different subset of items. Thus most available datasets for recommender systems naturally fit the requirements for pre-training. Connecting this to the general formulation of a contextual bandit problem, each arm $a_k$ is a particular item recommendation and user ids and/or any additional information constitutes the context $s$. Rewards $r \sim p_{a_t}$ in this context are ratings given by a user to recommended items. 

The decision making process in this case can be explicitly handled by finding a suitable acquisition function for recommender systems. While this choice most likely depends on the goals of a particular business, we will provide a proof-of-concept analysis, explicitly maximising coverage over the input space to provide the best-possible function approximation. This is motivated by the RMSE metric used in the literature on the experiments we consider. Inspired by work on decision trees, a natural criterion for evaluation is the information gain at a particular candidate item/arm. Writing $\mathbf{r}_{\setminus i}$ to denote the reward for each arm except $i$ in the target set $\mathcal{T}$ (likewise for $\mathbf{a}_{\setminus i}$) and suppressing dependence on context $s$ for clarity, we can thus define the information gain $\mathcal{IG}$ at arm $a_i$:

\begin{equation}
\label{eq:info_gain}
    \mathcal{IG}(a_i) := \mathcal{H}\big(p(\mathbf{r}_{\setminus i}|\mathbf{a}_{\setminus i}, \mathcal{C})\big) - \mathbb{E}_{\hat{r}_i \sim p(\mathbf{r}_{i}|a_{i}, \mathcal{C})} \big[\mathcal{H}\big(p(\mathbf{r}_{\setminus i}|\mathbf{a}_{\setminus i}, \mathcal{C} \cup \{a_i, \hat{r}_i\})\big)\big]
\end{equation}

Note that this involves using samples of the model's predictive distribution $\hat{r}_j$ at arm $a_j$ to estimate the entropy given an additional piece of information. Assuming $p(r_i| a_i, \mathcal{C})$ is a univariate normal, we arrive at an intuitive equation to determine the expected optimal next arm $a^*$ for evaluation (Details in the Appendix).

We apply NPs to the Movielens 100k \& much larger 20m datasets \citep{harper2016movielens}. While the specific format vary slightly, in both cases we face the basic problem of recommending movies to a user, given side-information such as the movie genre, certain user features (100k only) or user-provided tags (20m only). In both cases we closely follow the suggested experimental setup in \citep{chen2018federated} which only show results on the smaller 100k version. Importantly, 20\% of the users are explicitly withheld from the training dataset to test for few-shot adaptation. This is non-standard comparing to mainstream literature, which typically use a large fraction (70-90\%) of ratings from all users as the training set. However, we argue that withholding users allows for a more realistic evaluation, as this allows us to test the desirable few-shot learning setup.

This setup is particularly interesting for NPs, recalling their application at test time without gradient steps (see discussion in section \ref{sec:nps}). Thus, the model can be straight-forwardly deployed for a new user and cheaply run on-device as no backpropagation is required. Provided this works to a satisfactory degree, this property may be particularly desirable for fast recommendation on mobile devices. Finally, NPs can be trained in a federated learning setting, i.e. ratings provided by a user never need to leave the device as long as gradients can be shared, which may be an important advantage of this method when data privacy is of concern.

\begin{table}[]
\scriptsize
\setlength\tabcolsep{2pt}
\centering
\caption{Results on MovieLens 100k (\textbf{top}) and 20m (\textbf{bottom}). For both datasets, we report results given a varying percentage of ratings for \textit{new users} in the test set. Shown is the RMSE. Baseline results for 100k taken from \citep{chen2018federated}. Where provided we also report the standard deviations for 5 random splits of datapoints into observed and unobserved points given the same test set users. For SVD++, we also report whether results are obtained using only a test user's observed ratings (SELF) or all training set ratings but no ratings from the user under evaluation (MIXED). A discussion of this trade-off can be found in \cite{chen2018federated}.}
\label{table:movielens}
\begin{tabular}{@{}llll@{}}
\toprule
\textbf{}      & \textbf{}                                      & \textbf{MovieLens 100k}\\
\midrule
\textbf{Model} & \multicolumn{1}{l}{\textbf{20\% of user data}} & \textbf{50\%} & \textbf{80\%} \\ \midrule
SVD++ {\tiny(SELF)}          & 
$1.0517_{\color{white} \pm 0.0000}$ \tikz{\draw[gray,line width=.3pt] (0,0) -- (2,0);\draw[fill=black] (2.0,0) circle(2pt);} & 
$1.0217_{\color{white} \pm 0.0000}$ \tikz{\draw[gray,line width=.3pt] (0,0) -- (2,0);\draw[fill=black] (1.69387755102,0) circle(2pt);}&
$1.0124_{\color{white} \pm 0.0000}$ \tikz{\draw[gray,line width=.3pt] (0,0) -- (2,0);\draw[fill=black] (1.59897959184,0) circle(2pt);}           
\\
Multitask MLP  & 
$0.9831_{\color{white} \pm 0.0000}$ \tikz{\draw[gray,line width=.3pt] (0,0) -- (2,0);\draw[fill=black] (1.3,0) circle(2pt);}&
$0.9679_{\color{white} \pm 0.0000}$ \tikz{\draw[gray,line width=.3pt] (0,0) -- (2,0);\draw[fill=black] (1.14489795918,0) circle(2pt);}& 
$0.9507_{\color{white} \pm 0.0000}$ \tikz{\draw[gray,line width=.3pt] (0,0) -- (2,0);\draw[fill=black] (0.969387755102,0) circle(2pt);}
\\
MAML & 
$0.9593_{\color{white} \pm 0.0000}$ \tikz{\draw[gray,line width=.3pt] (0,0) -- (2,0);\draw[fill=black] (1.05714285714,0) circle(2pt);}& 
$0.9441_{\color{white} \pm 0.0000}$ \tikz{\draw[gray,line width=.3pt] (0,0) -- (2,0);\draw[fill=black] (0.902040816327,0) circle(2pt);}& 
$0.9295_{\color{white} \pm 0.0000}$ \tikz{\draw[gray,line width=.3pt] (0,0) -- (2,0);\draw[fill=black] (0.75306122449,0) circle(2pt);}
\\ 
\midrule
NP (random)  & 
$0.9359_{\pm 0.0017}$ \tikz{\draw[gray,line width=.3pt] (0,0) -- (2,0);\draw[fill=black] (0.818367346939,0) circle(2pt);}& 
$0.9215_{\pm 0.0055}$ \tikz{\draw[gray,line width=.3pt] (0,0) -- (2,0);\draw[fill=black] (0.671428571429,0) circle(2pt);}&
$0.9151_{\pm 0.0085}$ \tikz{\draw[gray,line width=.3pt] (0,0) -- (2,0);\draw[fill=black] (0.60612244898,0) circle(2pt);}                                    
\\
NP (Info gain) & 
$0.9288_{\pm 0.0011}$ \tikz{\draw[gray,line width=.3pt] (0,0) -- (2,0);\draw[fill=black] (0.745918367347,0) circle(2pt);}&
$0.8829_{\pm 0.0020}$ \tikz{\draw[gray,line width=.3pt] (0,0) -- (2,0);\draw[fill=black] (0.277551020408,0) circle(2pt);}&
$0.8557_{\pm 0.0084}$ \tikz{\draw[gray,line width=.3pt] (0,0) -- (2,0);\draw[fill=black] (0.0,0) circle(2pt);}
\\
\bottomrule
\end{tabular}
\vspace{0.3cm}
\\
\begin{tabular}{@{}llll@{}}
\toprule
\textbf{}      &               & \textbf{MovieLens 20m}  &               \\ 
\midrule
\textbf{Model} & \multicolumn{1}{l}{\textbf{20\% of user data}} & \textbf{50\%} & \textbf{80\%} \\ \midrule
SVD++ {\tiny(MIXED)}          & 
$0.9454_{\pm 0.0002}$ \tikz{\draw[gray,line width=.3pt] (0,0) -- (2,0);\draw[fill=black] (2.0,0) circle(2pt);} & 
$0.9454_{\pm 0.0005}$ \tikz{\draw[gray,line width=.3pt] (0,0) -- (2,0);\draw[fill=black] (2.0,0) circle(2pt);} & 
$0.9452_{\pm 0.0010}$ \tikz{\draw[gray,line width=.3pt] (0,0) -- (2,0);\draw[fill=black] (1.99845976126,0) circle(2pt);}
\\
Multitask MLP  & 
$0.8570_{\pm 0.0003}$ \tikz{\draw[gray,line width=.3pt] (0,0) -- (2,0);\draw[fill=black] (1.31921447824,0) circle(2pt);}& 
$0.8401_{\pm 0.0003}$ \tikz{\draw[gray,line width=.3pt] (0,0) -- (2,0);\draw[fill=black] (1.18906430497,0) circle(2pt);}& 
$0.8348_{\pm 0.0008}$ \tikz{\draw[gray,line width=.3pt] (0,0) -- (2,0);\draw[fill=black] (1.14824797844,0) circle(2pt);}       
\\
MAML          & 
$0.8142_{\pm 0.0043}$ \tikz{\draw[gray,line width=.3pt] (0,0) -- (2,0);\draw[fill=black] (0.989603388525,0) circle(2pt);} & 
$0.7852_{\pm 0.0019}$ \tikz{\draw[gray,line width=.3pt] (0,0) -- (2,0);\draw[fill=black] (0.76626877166,0) circle(2pt);} & 
$0.7780_{\pm 0.0048}$ \tikz{\draw[gray,line width=.3pt] (0,0) -- (2,0);\draw[fill=black] (0.710820177127,0) circle(2pt);}
\\ \midrule
NP (random)   & 
$0.7982_{\pm 0.0002}$ \tikz{\draw[gray,line width=.3pt] (0,0) -- (2,0);\draw[fill=black] (0.866384289565,0) circle(2pt);} & 
$0.7684_{\pm 0.0003}$ \tikz{\draw[gray,line width=.3pt] (0,0) -- (2,0);\draw[fill=black] (0.636888717751,0) circle(2pt);} & 
$0.7570_{\pm 0.0006}$ \tikz{\draw[gray,line width=.3pt] (0,0) -- (2,0);\draw[fill=black] (0.549095109742,0) circle(2pt);} \\
NP (Info gain) & 
$0.7926_{\pm 0.0003}$ \tikz{\draw[gray,line width=.3pt] (0,0) -- (2,0);\draw[fill=black] (0.82787832114,0) circle(2pt);} & 
$0.7362_{\pm 0.0005}$ \tikz{\draw[gray,line width=.3pt] (0,0) -- (2,0);\draw[fill=black] (0.391990758568,0) circle(2pt);} & 
$0.6859_{\pm 0.0006}$ \tikz{\draw[gray,line width=.3pt] (0,0) -- (2,0);\draw[fill=black] (0.0,0) circle(2pt);}       \\ 
\bottomrule
\end{tabular}

\end{table}

Results for random context sets of 20\%/50\%/80\% of test user's ratings are shown in Table \ref{table:movielens}. Thus, this shows the approximation error to the user's rating function given certain amounts of observed data. While these results are encouraging, the treatment as a decision making process using our acquisition function (denoted info. gain) leads to much stronger improvement. This indicates that the use of model uncertainty may be particularly appealing when new acquisition functions are designed.

Finally, we would like to mention that we expect that results may be further improved by also considering the model $f_i: u \mapsto r$, i.e. the item-specific function mapping from users $u$ to ratings $r$. Optimally, a hybrid model between both user-specific and item-specific function ought to be used. We leave this for future work.

\subsection{Adversarial Task search for RL agents}
\label{sec:adversarial_task_search}

As modern machine learning methods are approaching sufficient maturity to be applied in the real world, understanding failure cases of intelligent systems has become an important topic in our field, of paramount importance to efforts improving robustness and understanding of complex algorithms. One class of approaches towards identifying failure cases use adversarial attacks. The objective of an attack is to find a perturbation of the input such that predictions of the method being tested change dramatically in an unexpected fashion \cite[e.g.][]{szegedy2013intriguing, goodfellow2014explaining}.

Inspired by this recent line of work, we consider the recent study of \citep{ruderman2018uncovering} concerning failures of RL agents. The authors show that supposedly superhuman agents trained on simple navigation problems in 3D-mazes catastrophically fail when challenged with adversarially designed task instances trivially solvable by human players. The authors use an evolutionary search technique that modifies previous mazes based on the agent’s episode reward. A crucial limitation of this approach is that the evolution technique results in mazes with zero probability under the task-distribution, weakening the significance of the results.

Instead, we propose to tackle the worst-case search through a Bayesian Optimisation approach on a fixed set of possible candidate mazes using a Neural Process surrogate model. More formally, we study the adversarial task search problem on mazes as follows: Given an (unobserved) agent $A$ under adversarial test, parameters of the task at hand (some maze layout $M$, start and goal positions $p_{s}, p_{g}$ defining the navigation problem), we can think of the agent as some function $f_A$, mapping from task parameters to its performance $r$ (implicitly through the agent's policy). Thus, two  Bayesian optimisation problems emerge for adversarial testing:

\begin{enumerate}
    \item[(i)] \textbf{Position search:} The search for the most difficult start and goal position within a fixed maze layout: $p_s^*, p_g^* = \argmin_{p_{s}, p_{g}} f_A(M, p_{s}, p_{g})$
    \item[(ii)] \textbf{Full maze search:} The search for the most difficult navigation problem, including maze layout, start and goal positions: $p_s^*, p_g^*, M^* = \argmin_{p_{s}, p_{g}, M} f_A(M, p_{s}, p_{g})$
\end{enumerate}

Note that the complexity of problem (ii) quickly becomes prohibitive, as evaluation of all possible candiatie location scales as $\mathcal{O}(NKC)$ for $N$ BO iterations, a total set of $K$ available maze layouts and $C$ possible pairs of start and goal location within each maze. However, by reusing an existing BO solution to problem (i), we can reduce the complexity to $\mathcal{O}(\frac{N}{l} (K + lC))$, where we allow for $l$ iterations of BO on problem (i).\footnote{One can decompose the problem into \textbf{A}: Finding mazes that are likely to be difficult and \textbf{B}: Identifying the most difficult start and goal locations within this maze (reusing the solution to problem (i)).}

\begin{figure*}[t!]
    \centering
    \subfigure[Position search results]{\label{fig:position_model_results}\includegraphics[width=0.45\linewidth]{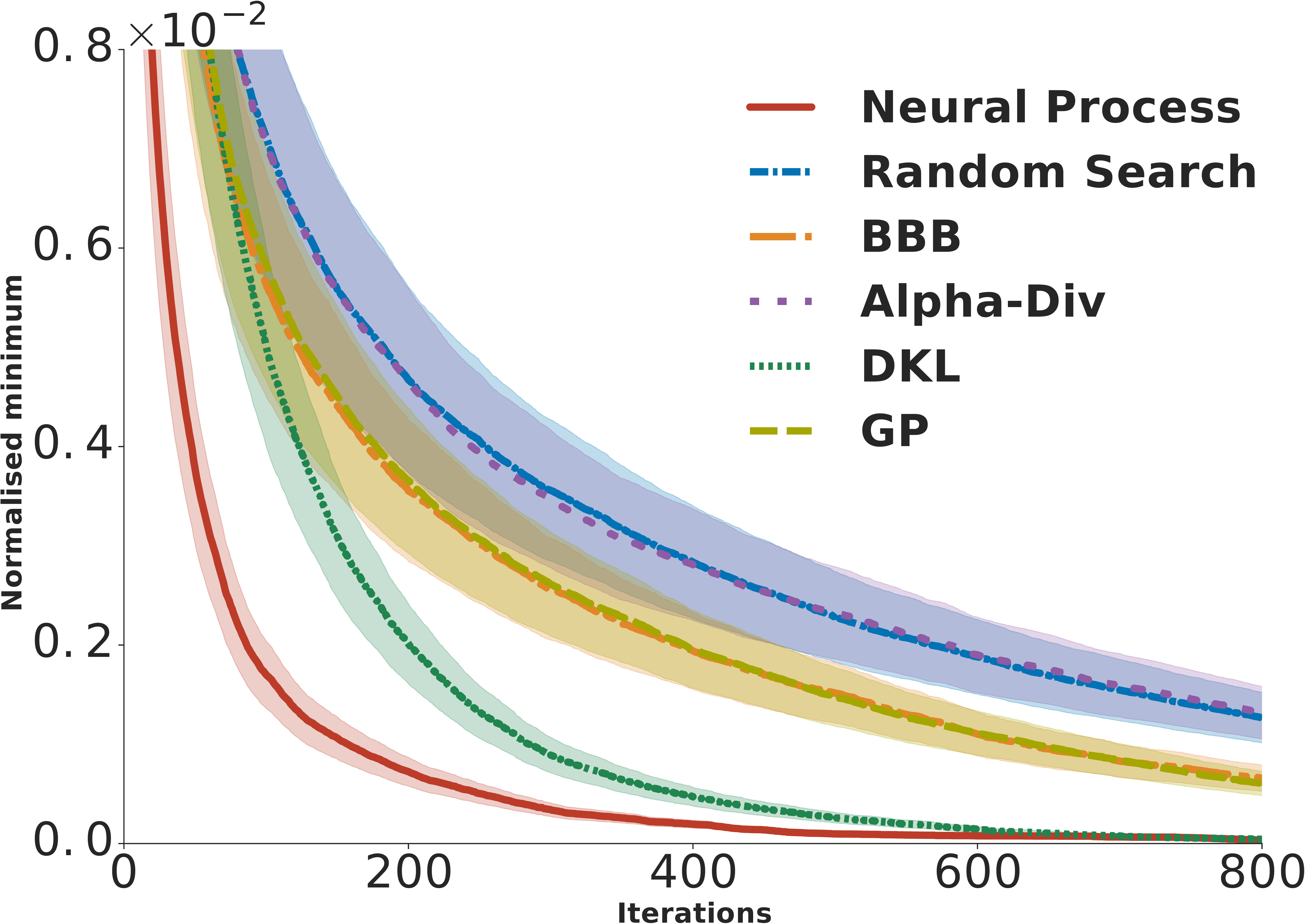}}
    \subfigure[Full maze search results]{\label{fig:full_results}\includegraphics[width=0.45\linewidth]{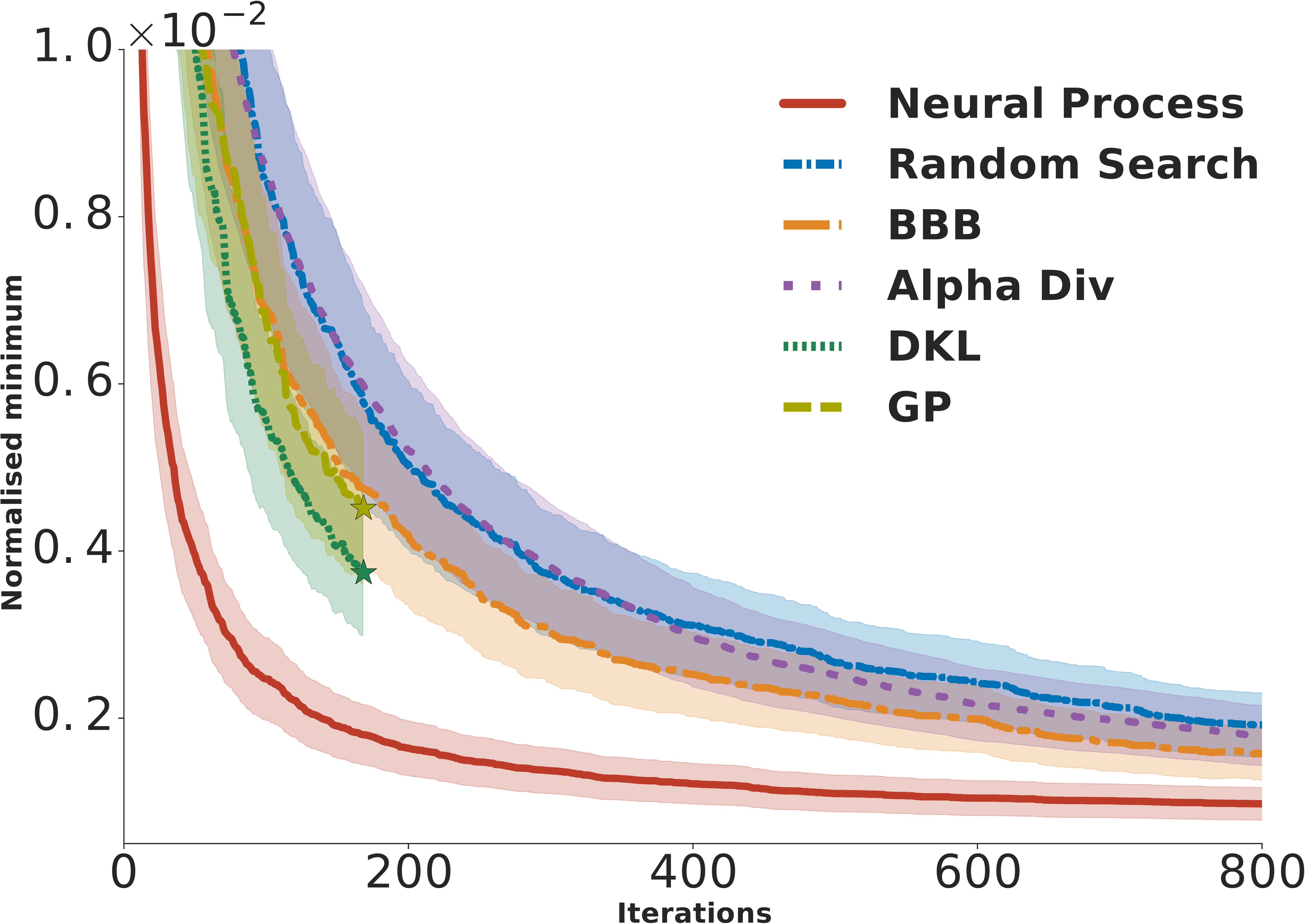}}
    \caption{Bayesian Optimisation results. \textbf{Left:} Position search \textbf{Right:} Full maze search. We report the minimum up to iteration $t$ (scaled in [0,1]) as a function of the number of iterations. Bold lines show the mean performance over 4 unseen agents on a set of held-out mazes. We also show 20\% of the standard deviation. \textbf{Baselines:} GP: Gaussian Process (with a linear and Matern 3/2 product kernel~\citep{NIPS2008_multitask_gp}), BBB: Bayes by Backprop \citep{blundell2015weight}, AlphaDiv: AlphaDivergence \citep{hernandez2016black}, DKL: Deep Kernel Learning \citep{wilson2016deep}.}
\end{figure*}

Addressing first the question of performance on the \textit{position search} problem, we show results in Figure~\ref{fig:position_model_results} indicating strong performance for NPs when applied in our framework. Indeed, we find start and goal positions close to the minimum after evaluating only approx. 5\% of the possible search space. Most iterations are spent on determining the global minimum among a relatively large number of goal positions of similar return magnitude. In practice, if a point close enough to the minimum is sufficient to determine the existence of an adversarial maze, search can be terminated much earlier. In order to account for pretraining, we reuse embeddings of the inputs obtained from the NP for all baselines, which significantly improves performance. 

\begin{wrapfigure}{L}{0.5\textwidth}
\vspace{-15pt}
\begin{minipage}{0.5\textwidth}
    \begin{figure}[H]
        \centering
        \subfigure[Few shot predictions]{\label{fig:pos_model_analysis_a}\includegraphics[width=\linewidth]{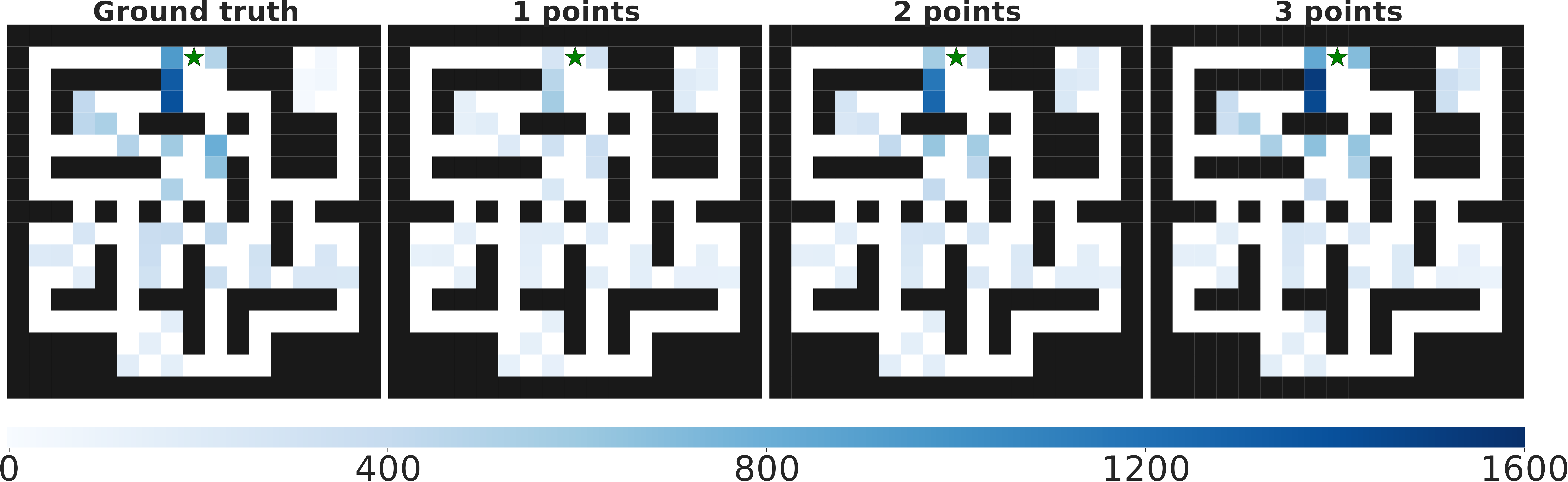}}
        
        \subfigure[Uncertainty analysis]{\label{fig:pos_model_analysis_b}\includegraphics[width=0.9\linewidth]{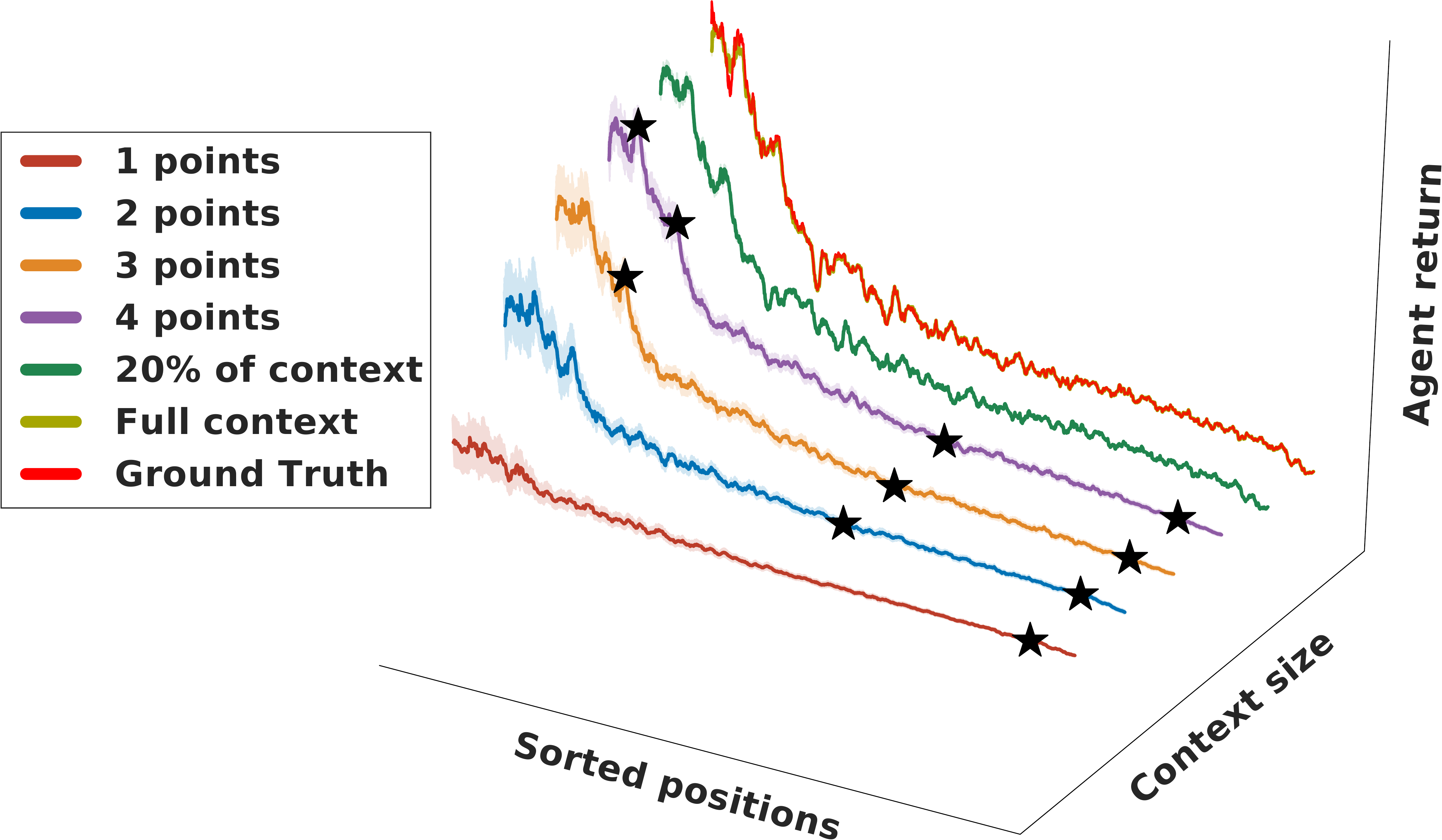}}
        
        \subfigure[Worst examples trajectories]{\label{fig:bad_example_maps}\includegraphics[width=\linewidth]{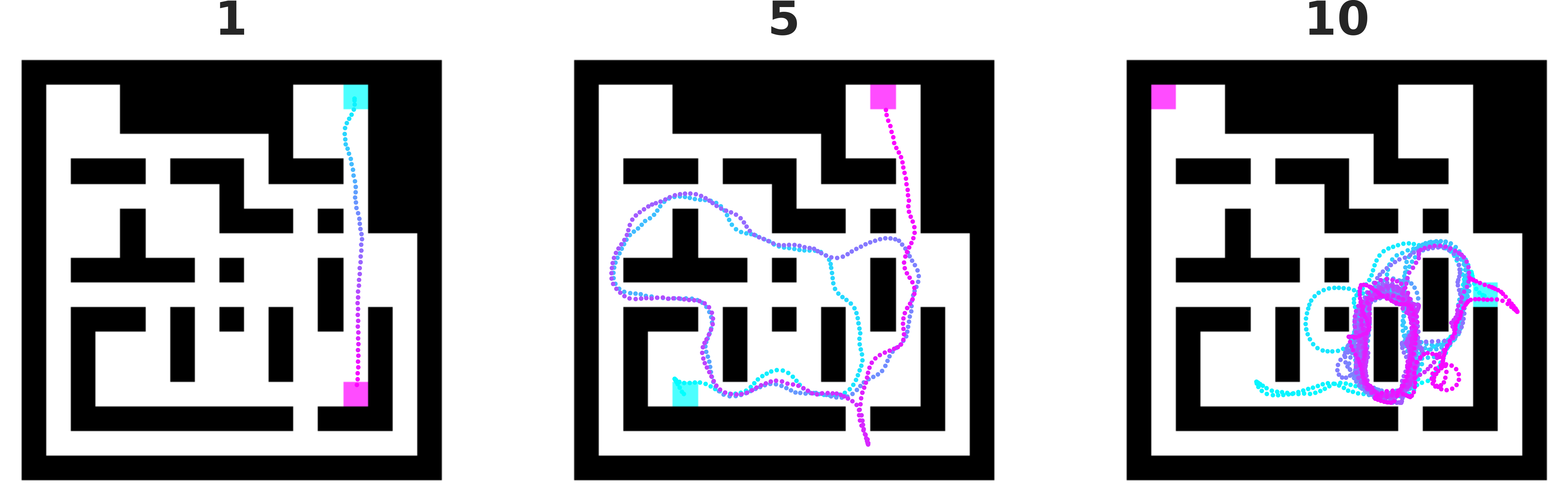}}
        
        \caption{\textbf{Top:} Agent returns for a fixed start (green star) and varying goal positions (blue squares). Colour intensity indicates return magnitude. Shown is the ground-truth and few-shot predictions. \textbf{Middle:} Predicted episode returns and model uncertainty of a NP. Positions are sorted by the absolute distance between start and goal positions. Stars indicate observed datapoints. \textbf{Bottom:} Trajectories of an agent after 1, 5 and 10 position-model optimisation iterations. The procedure successfully results in failure of the agent to solve the navigation task. The blue and magenta rectangles indicates the start and goal position respectively.}
    \end{figure}
\end{minipage}
\vspace{-40pt}
\end{wrapfigure}

In order to explain the sources of this improvement, we show an analysis of Neural Process uncertainty in function space in Figure \ref{fig:pos_model_analysis_b} for varying context sizes. The graphic should be understood as the equivalent of Figure 3 in \citep{garnelo2018neural} for the adversarial task search problem. More specifically, we plot functions drawn from a neural process (i.e. predictions on all possible goal positions for a given start position) given varying number of observed points (shown as stars)). As expected, a small number of positions near the start location result in high returns, while positions further away result in lower returns.

As we would expect, uncertainty in function space decreases significantly as additional context points are introduced. Furthermore, we observe an interesting change in predictions once context points two and three are introduced (blue and orange lines). The mean prediction of the model increases noticeably, indicating that the agent being evaluated performs superior to the mean agent encountered during pre-training. The model quickly adapts to the higher scale of the rewards.

Showing the advantage of pre-training our method, we illustrate episode reward predictions on mazes given small context sets in Figure \ref{fig:pos_model_analysis_a}. Note that the model has learned to assign higher scores to points closer to the starting location taking into account obstacles, without any explicitly defined distance metric provided. Thus, predictions incorporate this valuable prior information and are fairly close to the ground truth after a single observation.

Finally, we test our model in full case search on holdout mazes, using the proposed approach for the full maze search to reduce search complexity as outlined above. From Figure~\ref{fig:full_results}, we continue to observe superior performance for this significantly more difficult problem. A video demonstrating a successful run of the final optimiser corresponding to the results in Figure \ref{fig:bad_example_maps} can be found here: \url{https://tinyurl.com/y47ccef9}.

\section{Related work}
\label{sec:related_work}

There has been a recent surge of interest in Meta-Learning or Learning to Learn, resulting in large array of methods \citep[e.g.][]{koch2015siamese, andrychowicz2016learning, wang2016learning}, many of which may be applied in the problems we study (as we merely assume the existence of a general method for regression). 

Nevertheless, several recent publications focus on probabilistic ideas or re-interpretations \citep[e.g.][]{bauer2017discriminative, bachman2018vfunc}, and are thus applicable in our framework. An example is Probabilistic MAML \citep{finn2018probmodel} which forms an extension of the popular model-agnostic Meta-Learning (MAML) algorithm \citep{finn2017model} that can be learned with variational inference. Other recent works cast Meta-Learning as hierarchical Bayesian inference \citep[e.g.][]{edwards2016towards, hewitt2018variational, grant2018recasting, ravi2018amortized}.

Gaussian Processes (GPs) are popular candidates due to closed-form Bayesian inference and have been used for several of the problems we study \citep[e.g.][]{krause2011contextual, deisenroth2011pilco, saemundsson2018meta}. While providing excellent uncertainty estimates, the scale of modern datasets can make their application difficult, thus often requiring approximations \citep[e.g.][]{titsias2009variational}. Furthermore, their performance strongly depends on the choice of the most suitable kernel (and thus prior over function), which may in practice require careful design or kernel learning techniques \citep[e.g.][]{duvenaud2013structure, wilson2016deep}. Moreover, much of the recent work on Bayesian Neural Networks \citep[e.g.][]{blundell2015weight, gal2016dropout, hernandez2016black,  louizos2017multiplicative} serves as a reasonable alternative, also benefiting from the flexibility and power of modern deep learning architectures.

Finally, the approach in \citep{chen2017learning} tackles similar problems, applying Meta-Learning for black-box optimisation, directly suggesting the next point for evaluation. 

\label{sec:discussion}

In this paper, we introduced a general framework for applying Meta-Learning techniques to decision making problems, showing competitive results over a broad range of domains. At this point we would like to remind the reader that no aspect of the models used in the experiments has been tailored towards the problems we study. Our RL experiments showed significant improvements in terms of data efficiency when a large set of related tasks is available. In future work, it would be interesting to consider more complex problems, which may require a more sophisticated policy during pre-training. The presented results for recommender systems in particular are encouraging, noting that many of the standard tricks used for such systems are orthogonal to NPs and could thus be easily incorporated. Our experiments on adversarial task search indicate that such a system may for instance be used within an agent evaluation pipeline to test for exploits. A possible avenue of future work could utilise the presented method to train more robust agents, suggesting problems the agent is currently unable to solve.

\newpage
\section{Appendix}
\subsection{Algorithmic details}
\subsubsection{Neural Processes}

\label{sec:nps}
Given a number of realisations from some unknown stochastic process $f:\mathcal{X} \to \mathcal{Y}$, NPs can be used to predict the values of $f$ at some new, unobserved locations. In contrast to standard approaches to supervised learning such as linear regression or standard neural networks, NPs model a distribution over functions that agree with the observations provided so far (similar to e.g. Gaussian Processes \citep{rasmussen2003gaussian}). This is reflected in how NPs are trained: We require a dataset of evaluations of similar functions $f_1, \dotsm f_n$ over the same spaces $\mathcal{X}$ and $\mathcal{Y}$. Note however, that we do not assume each function to be evaluated at the same $x \in \mathcal{X}$. Examples of such datasets could be the temperature profile over a day in different cities around the world or evaluations of functions generated from a Gaussian process with a fixed kernel. 
In order to allow NPs to learn distributions over functions, we split evaluations $(x_i, y_i)$ for each function $f$ into two disjoint subsets: a set of $m$ context points $\mathcal{C} = \{(x_i, y_i)\}_{i=1}^m$ and a set of targets $\mathcal{T} = \{(x_j, y_j)\}_{j=m+1}^{n}$ that contains $n-m$ unobserved points. These data points are then processed by a neural network as follows:
\begin{align}
    r_i &= h_{\theta}(x_i,y_i) \qquad \forall (x_i, y_i) \in \mathcal{C}\\
    r &= r_1 \oplus r_2 \oplus \ldots r_{n-1} \oplus r_n \label{eq:embedding}\\ 
    z & \sim \mathcal{N}(\mu_{\psi_1}(r, \log(n)), \sigma_{\psi_2}(r, \log(n))) \\
    \phi_i &= g_{\omega}(x_j, z) \qquad \forall (x_j) \in \mathcal{T} \label{eq:np_decoder}
\end{align}

First, we use an encoder $h_{\theta}$ with parameters, transforming all $(x_i, y_i)$ in the context set to obtain representations $r_i$. We then aggregate all $r_i$ to a single representation $r$ using a permutation invariant operator $\oplus$ (such as addition) that captures the information about the underlying function provided by the context points. Later on, we parameterise a distribution over a latent variable $z$, here assumed to be Normal with $\mu, \sigma$ estimated by an encoder network using parameters $\psi_{1},\psi_{2}$. Note that this latent variable is introduced to model uncertainty in function space, extending the Conditional Neural Process \citep{garnelo2018conditional}. 

Thereafter, a decoder $g_{\omega}$ is used to obtain predictive distributions at target positions $x_i \in \mathcal{T}$. Specifically, we have $p(y_i| x_i, z; \phi_i)$ with parameters $\phi_i$ depending on the data modelled. In practice, we might decide to share parameters, e.g. by setting $\theta \subset \omega$ or $\psi_1 \cap \psi_2 \neq \emptyset$. To reduce notational clutter, we suppress dependencies on parameters from now on.

In order to learn the resulting intractable objective, approximate inference techniques such as variational inference are used, leading to the following evidence lower-bound:

\begin{footnotesize}
\begin{equation}
\label{eq:elbo1}
\begin{split}
    &\log p(y_{m+1:n}|x_{1:n},y_{1:m}) \\
    \ge & \mathbb{E}_{q(z|x_{1:n},y_{1:n})} \left[
    \sum_{i=m+1}^n \log p(y_i|z,x_i) + \log\frac{p(z|x_{1:m},y_{1:m})}{q(z|x_{1:n},y_{1:n})} 
    \right]
\end{split}
\end{equation}
\end{footnotesize}

which is optimised with mini-batch stochastic gradient descent using a different function $f_j$ for each element in the batch and sampling $|\mathcal{C}|, |\mathcal{T}|$ at each iteration.

Recently, attention has been successfully applied for NPs \citep{kim2018attentive}, improving predictions at observed points. For various alternatives to the loss in \eqref{eq:elbo1} we refer the interested reader to~\citep{anh2018empirical}. 

\subsubsection{Neural Processes for Bayesian Optimisation}

\begin{algorithm}[tb]
  \caption{Bayesian Optimisation with NPs and Thompson sampling.}
  \label{alg:bo_np}
\begin{algorithmic}
  \STATE {\bfseries Input:}
  \STATE $f$ - Function to evaluate
  \STATE $\mathcal{C}_{0} = \{(x_{0}, y_{0})\}$ - Initial randomly drawn context set
  \STATE $N$ - Maximum number of function iterations
  \STATE $\mathcal{NP}$ - Neural process pre-trained on evaluations of similar functions $f_1, \dots f_n$
  \STATE
  \FOR{n=1, \ldots, N}
  \STATE Infer conditional $\mathcal{NP}$ prior $q(z|{C}_{n-1})$
  \STATE Thompson sampling: Draw $z_{n} \sim q(z|{C}_{n-1})$, find
  \STATE \begin{equation} x_{n} = \argmin_{x \in \mathcal{X}} \mathbb{E} \big[g(y|x, z_{n})\big] \label{eq:thompson_sampling}\end{equation}
  \STATE 
  \STATE Evaluate target function and add result to context set
  \STATE $\mathcal{C}_n \leftarrow \mathcal{C}_{n-1} \cup \{(x_{n}, f(x_{n}))\}$
  \ENDFOR
\end{algorithmic}
\end{algorithm}

A more specific description of the Baysian Optimisation algorithm with NPs as the surrogate model of choice is shown in Algorithm \ref{alg:bo_np}, using notation introduced in the previous subsection. Note the absence of the model adaptation step in comparison to the more general formulation. Thus, NPs have the appeal of fast inference (see discussion on Recommender Systems) and the lack of any hyper-parameters that control the adaption behaviour at test time.

\subsection{Model-based RL}
\label{appendix:model_based}

\begin{algorithm}
  \caption{Meta-learning for Model-based Reinforcement Learning}
  \label{alg:mbrl}
\begin{algorithmic}
  \STATE {\bfseries Meta-training Input:}
  \STATE $\mathcal{M}_{\theta}$- Meta-learned surrogate model parameters $\theta$ to estimate $p: \mathcal{S} \times \mathcal{A} \rightarrow \mathcal{S}$ and $r: \mathcal{S}\times \mathcal{A} \rightarrow \mathcal{R}$
  \STATE $p(\mathcal{T})$ - Task Distribution
  \STATE $\pi_\phi$ - Exploratory policy.
  \STATE 
  \WHILE {pre-training is not finished}
    \STATE Sample a task $\mathcal{T}_w \sim p(\mathcal{T})$.
    \STATE Obtain transitions \{s, a, r, s'\} by acting with $\pi_\phi$ on $\mathcal{T}_w$.
    \STATE \texttt{// Model improvement}
    \STATE Optimise $\theta$ to improve $\mathcal{M}$'s prediction on $\mathcal{D}$.
  \ENDWHILE

  \STATE 
  \STATE {\bfseries Meta-testing Input:}
  \STATE $\psi$ - Policy parameters for target task $\mathcal{T}^*$, $\mathcal{R}$ - Replay
  \STATE $K$ - Rollout length, $M/N$ - Model/Policy training steps
  \STATE 
  \WHILE {true}
  \STATE Run $\pi_{\psi}$ on the real environment $\mathcal{T}^*$, obtain trajectory $\mathcal{\tau} = (s_1, a_1, r_1, \dots, s_K)$.
  \STATE $\mathcal{R} \leftarrow \mathcal{R} \cup \{\tau\}$
  \STATE
  \STATE \texttt{// Model-adaptation}
  \FOR{i=1,\ldots, M}
  \STATE Optimise $\theta$ to improve $\mathcal{M}$'s prediction on $\mathcal{R}$.
  \ENDFOR
  \STATE
  \STATE \texttt{// Policy learning}
  \FOR{j=1,\ldots, N}
  \STATE Sample an initial state $s_1 \sim \mathcal{R}$ observed on $w^*$.
  \STATE Generate trajectory $\tau'=\{a_1, r_1 \dots, s_k, a_k, r_k\}$ using $\pi_\psi$ and autoregressive sampling from $\mathcal{M}$.
  \STATE
  \STATE Update policy $\pi_{\psi}$ using $\tau'$ and any RL algorithm.
  \ENDFOR
  \ENDWHILE
\end{algorithmic}
\end{algorithm}

We provide a more concrete algorithmic description of our framework for Model-based Reinforcement in Algorithm \ref{alg:mbrl}, explicitly distinguishing between the necessary pre-training step for Meta-Learning methods and the few-shot adaptation at test time.

\subsubsection{Model-based Baselines}
\begin{itemize}
    \item \textbf{MAML}: The model-agnostic meta-learning algorithm \citep{finn2017model}. We use code provided by the authors, manually choosing the inner learning rate (as opposed to learning it). We apply the method to meta-learn the model as opposed to the policy, which is the common use-case.
    \item \textbf{Multitask MLP}: A straight-forward application of a MLP without accounting for the fact that data comes from separate tasks. Thus, the algorithm may receive identical inputs $x=(s, a)$ resulting in different transitions $y_1=(s', r) \neq y_2=(s', r)$ for tasks $\mathcal{T}_1$ and $\mathcal{T}_2$. Thus, the Multitask MLP baseline can at best learn the mean over all transition dynamics, therefore relying on optimisation to adjust from a suboptimal model to accurate model for task $\mathcal{T}^*$. This is similar to MAML with the caveat that the initialisation has not been optimisation for fast adaptation.
\end{itemize}

\begin{figure}
    \centering
    \subfigure[Neural Process (Jobs terminated early)]{\includegraphics[width=0.3\linewidth]{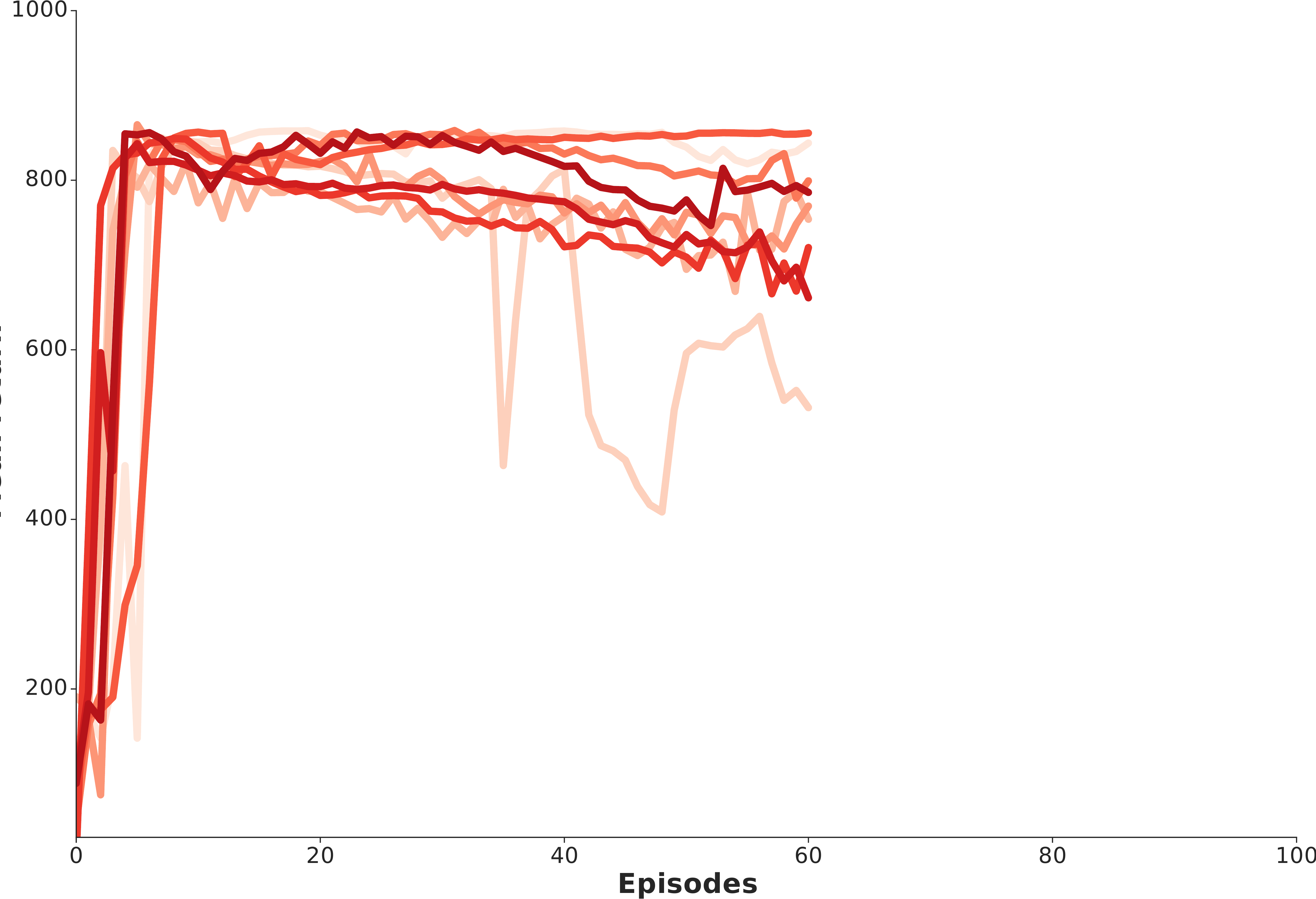}}
    \subfigure[MAML]{\includegraphics[width=0.3\linewidth]{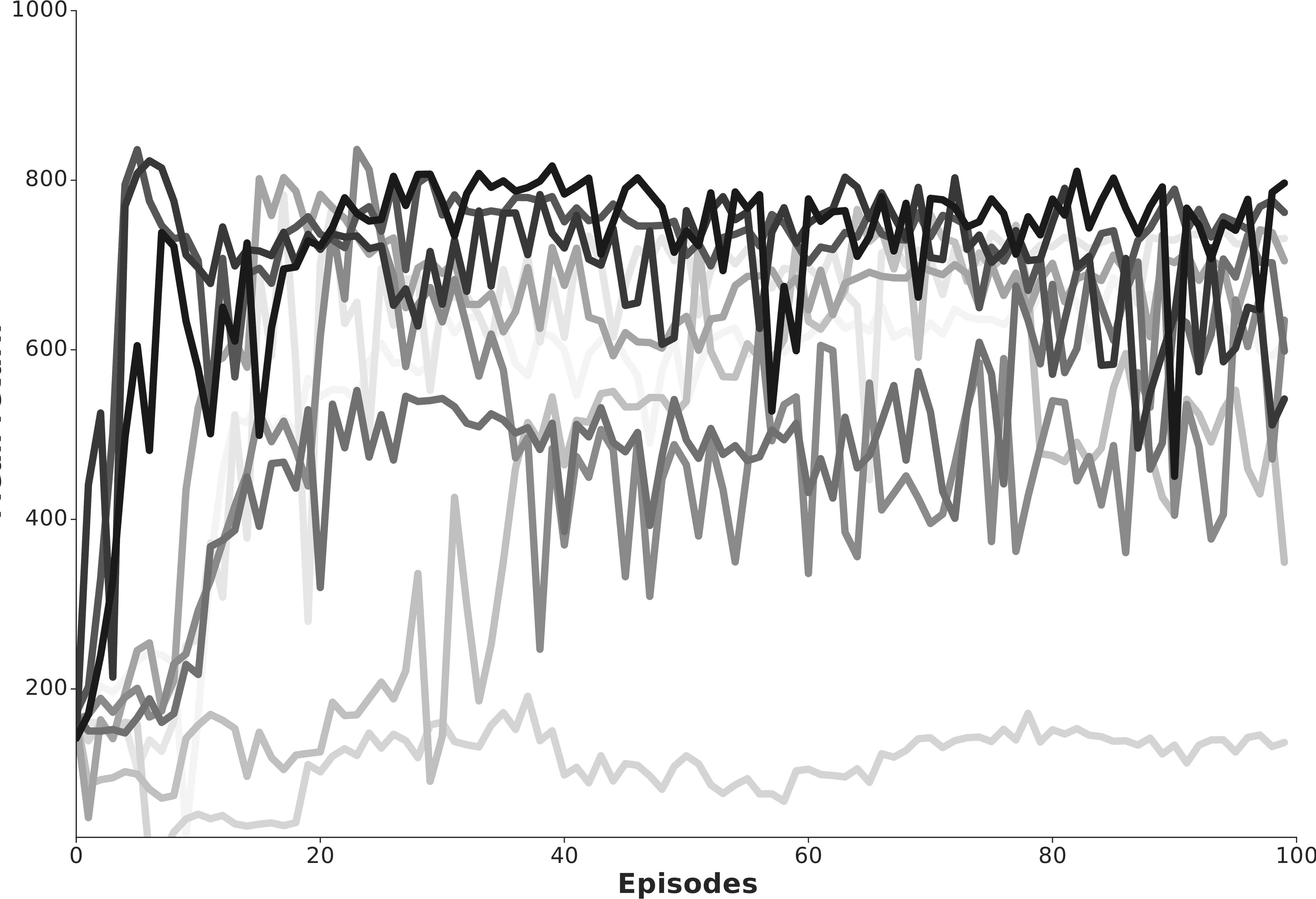}}
    \subfigure[Multitask MLP]{\includegraphics[width=0.3\linewidth]{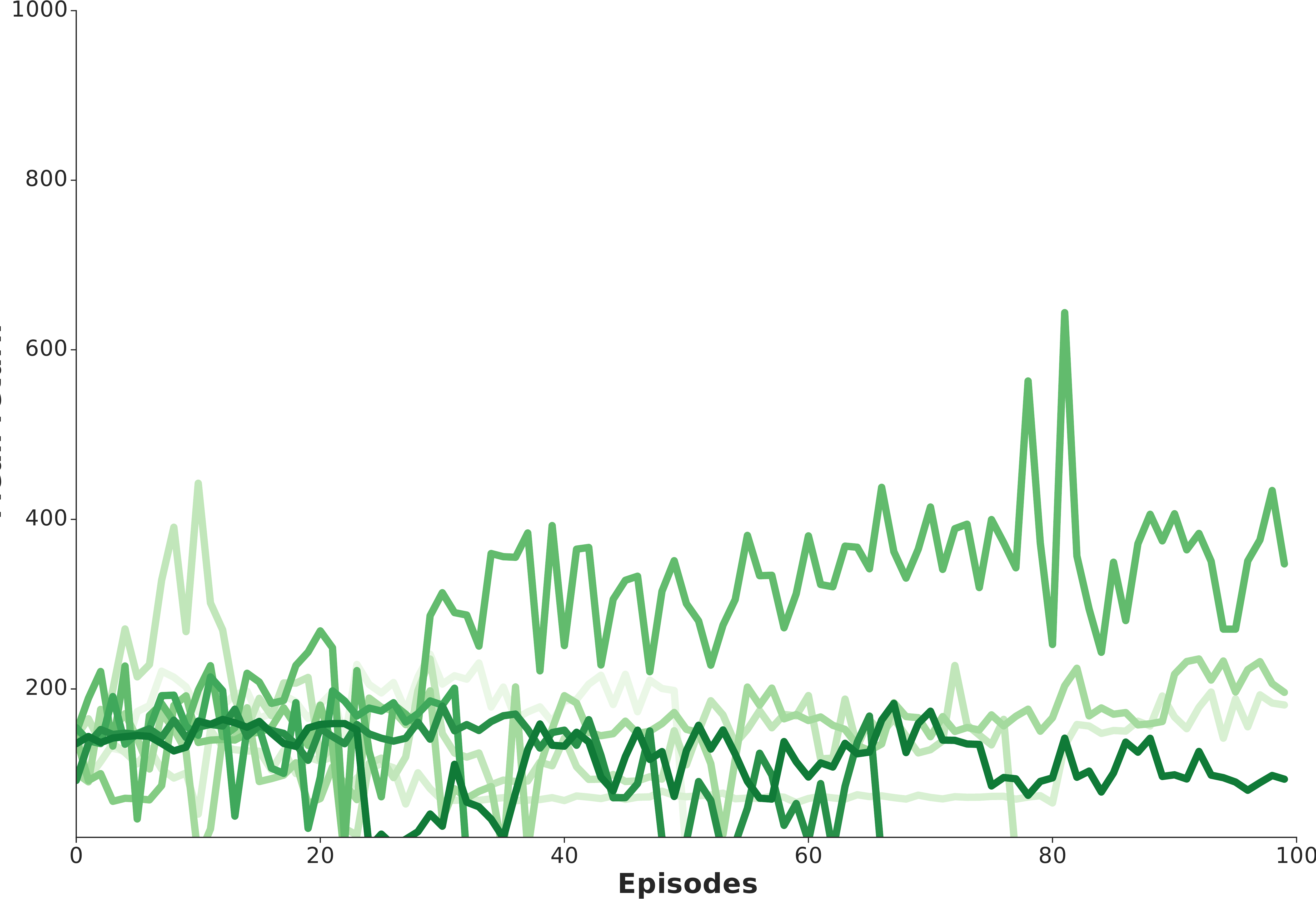}}
    \subfigure[D4PG]{\includegraphics[width=0.3\linewidth]{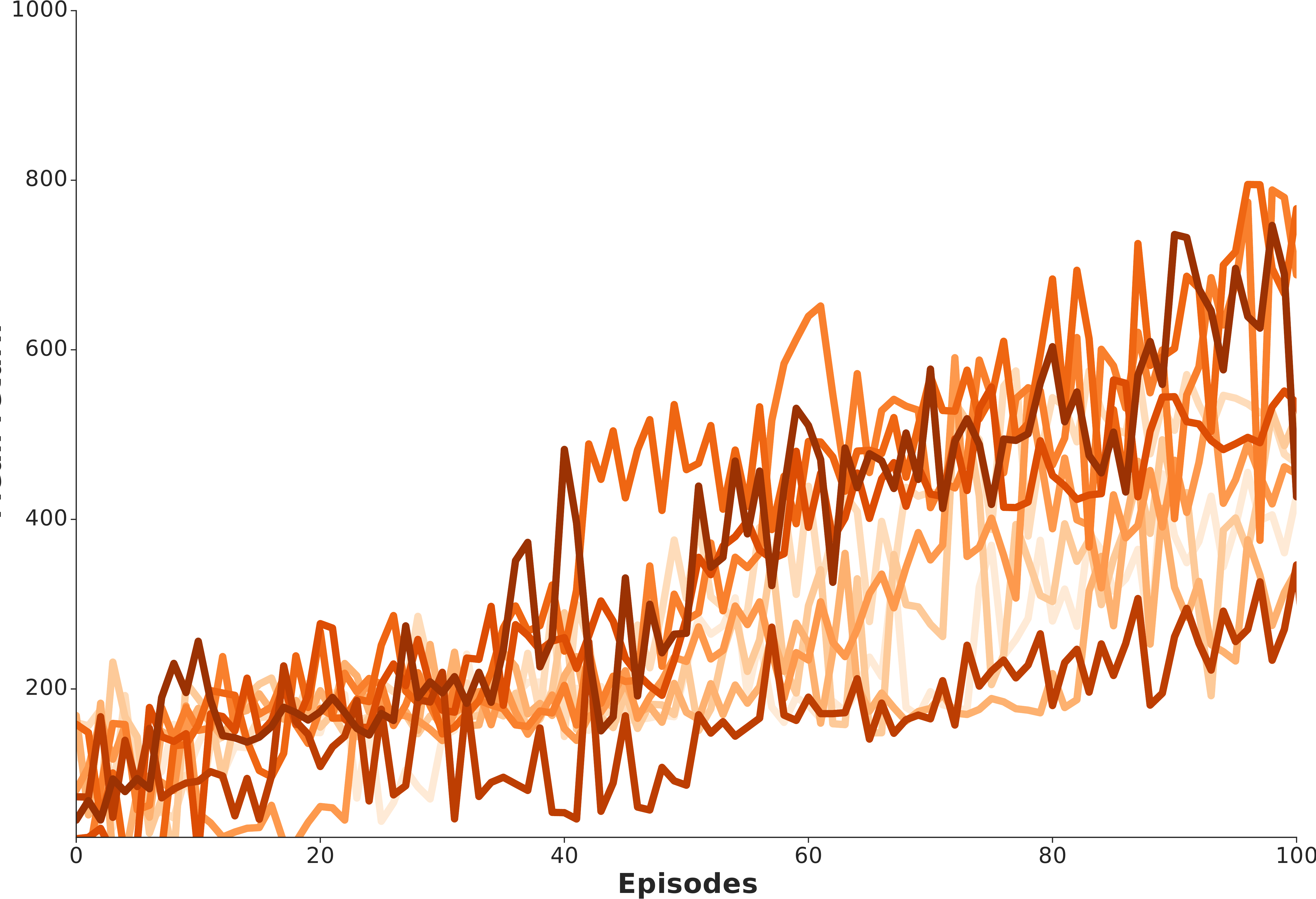}}
    \subfigure[SVG(0) + Retrace]{\includegraphics[width=0.3\linewidth]{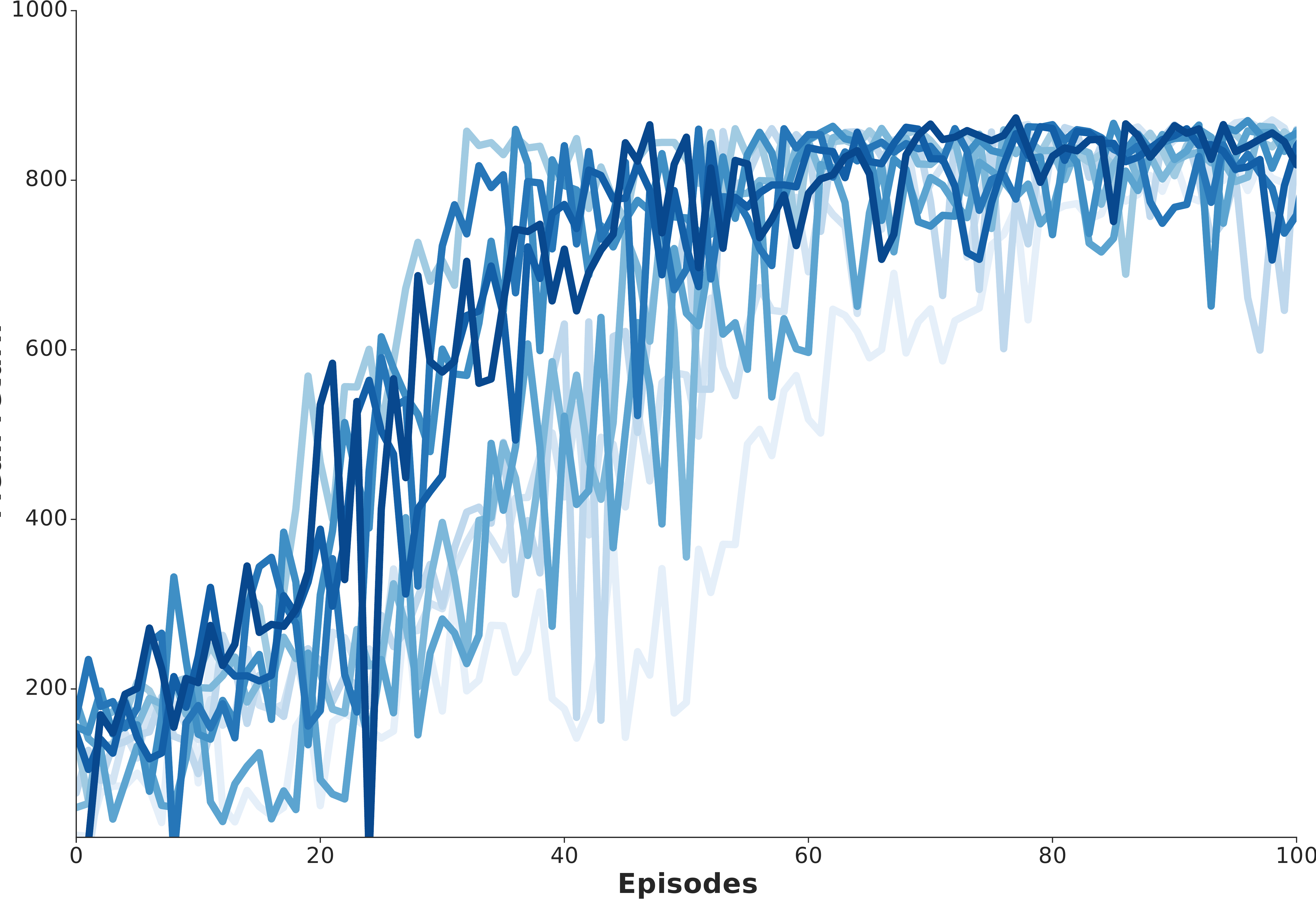}}    
    \caption{All runs for model-based and model-free RL methods.}
    \label{fig:runs_model_based_free}
\end{figure}

All model-based methods are pre-trained using an exploration policy $\pi_{\phi}$ to obtain transitions on related tasks. To minimise domain-knowledge, we use the following random walk:
\begin{equation}
    \label{eq:exploration_policy}
    a_{t} = sin(a_{0} + u \sum_{k=1}^{t} w_{k}),
\end{equation}
where $a_{0} \sim \mathcal{U} [0, 2 \pi]$, $u \sim \mathcal{U}[0, 1]$ are fixed for the entire episode and $w_{k} \sim \mathcal{N}(0,1)$. We show a video of this behaviour in \url{https://tinyurl.com/y5qf7j2v}.

An interesting option for future work would be to use a fixed amount of trajectories on related task from some expert policy. This is a realistic setup in robotics,
where imitation learning is a popular strategy \citep[e.g.][]{bakker1996robot, mendonca2019guided} and could be provided by a human expert. We assume this to be particularly beneficial in sparse reward tasks. In case no expert data is available and an appropriate hand-designed exploratory policy is difficult to obtain, one option would be to implement an efficient curiosity-driven algorithm. 

\subsubsection{Experimental details}

\begin{table}[]
\footnotesize
\centering
\caption{Hyperparameters for the Neural Process pretraining in model-based RL. Attention \citep{kim2018attentive} indicates whether an attentive neural process was used. Encoder and decoder indicate the MLP network sizes used. Context size $|\mathcal{C}|$ indicate the number of context points used during the training. $\mathcal{C} \subset \mathcal{T}$ denotes whether context datapoints are included in the target set. We also introduce a minimum value for the standard deviation of predictions at target points.
}
\label{table:model_based_rl_np_params}

\begin{tabular}{@{}llllll@{}}
\toprule
\textbf{} & \textbf{\textbf{Parameter}}           & \textbf{Considered Range}             & \textbf{MAML}         & \textbf{Multitask MLP} & \textbf{Neural Process} \\ \midrule
          & Encoder (\#Layers $\times$ Units)     & $\{2, 3\}\times\{128\}$      & $2\times128$ & $2\times128$  & $2\times128$   \\
          & Decoder (\#Layers $\times$ Units)     & $\{2, 3\}\times\{128\}$      & $2\times128$ & $2\times128$  & $2\times128$   \\
          & Meta Learning rate                    & \{$10^{-4}, 10^{-5}\}$       & $10^{-4}$    & $10^{-4}$     & $10^{-4}$      \\
          & Update Learning rate                  & \{$10^{-2}, 10^{-3}\}$       &              &               &                \\
          & Inner gradient steps                  & \{1, 5, 10\}                 & 10           &               &                \\
          & Number of training tasks              & \{2000\}                     & 2000         & 2000          & 2000           \\
          & Rollouts per training task            & \{10\}                       & 10           & 10            & 10             \\
          & Rollout length                        & {50, 100, 200}               & 100          & 100           & 100            \\ \midrule
          & Attention                             & \{None, Laplace, Multihead\} &              &               & Laplace        \\
          & $\mathcal{C} \in \mathcal{T}$         & \{True\}                     &              &               & True           \\
          & min $\sigma$                          & \{0.1\}                      &              &               & 0.1            \\
          & Fixed sigma                           & \{False\}                    &              &               & False          \\
          & Maximum context size: $|\mathcal{C}|$ & $\{50, 100, 200, 300, 350\}$ &              &               & 300            \\ \bottomrule
\end{tabular}
\end{table}

We show hyperparameters for this pre-training step in Table~\ref{table:model_based_rl_np_params}. For each hyperparameter, we run training for $10^6$ iterations, where at each iteration we apply learning updates for the batch of environment transitions. Each batch element corresponds to transitions sampled from a particular task.

Once pre-training has converged, we apply the meta-learned model and compare to model-free ones. Experimental details are shown in Table \ref{table:rl_params}. Results are reported with 10 random seeds for all methods. We report the mean episode reward over 50 test episodes using the current parameters and calculate the standard deviation across different simulations. For the results reported over a grid of held-out taks parameters (surface plot), we show the average across 5 seeds on a particular task instance at convergence. 

The model-free baselines shown in the experiments in the main text are SVG(0) \citep{heess2015learning} with Retrace off-policy correction~\citep{MunosSHB16} and D4PG \citep{barth2018distributed}. Finally, in the model-based RL literature \citep[e.g.][]{sutton2018reinforcement} it is not uncommon to use both real environment trajectories in addition to trajectories according to the model for policy learning. While this has not been considered in the experiments (as we merely evaluate up to 100 environment rollouts), this can be straight-forwardly done.

\begin{table}[]

\scriptsize
\centering
\caption{Hyperparameters for the experiments in model-based RL on cartpole. Shown are both the range of considered value as well as the best option for each model. For model-free baselines RS0 and D4PG, we make use of target networks which are updated every "target period update" learning steps.}
\label{table:rl_params}

\begin{tabular}{@{}lllllll@{}}
\toprule
\textbf{Parameter}                                     & \textbf{Considered Range}              & \textbf{RS0}       & \textbf{D4PG}      & \textbf{NP}                 & \textbf{M.MLP}              & \textbf{MAML}               \\ \midrule
Training steps per episode                             & $\{100, 500, 1000, 5000\}$             & $500$              & $100$              & $500$              & $1000$             & $1000$             \\
Batch size                                             & $\{32, 128, 512\}$                     & $512$              & $512$              & $128$              & $128$              & $128$              \\
Target period update                                   & $\{10, 50, 100\}$                      & $100$              & $100$              &                    &                    &                   \\
Entropy cost                                           & $\{10^{-2}, 10^{-3}, 10^{-4}\}$        & $10^{-2}$          & $10^{-2}$          & $10^{-3}$          & $10^{-3}$          & $10^{-3}$          \\
min $\sigma$                                           & $\{10^{-1}, 10^{-2}, 10^{-3}\}$        & $0.01$             & $0.01$             & $0.01$             & $0.01$             & $0.01$             \\
max $\sigma$                                           & $\{None, 0.6, 1.0\}$                   & None               & None               & $0.6$              & $0.6$              & $0.6$              \\
Rollout length                                         & $\{10, 50, 100, 150, 200\}$            & $10$               & $10$               & $100$              & $100$              & $100$              \\
Context size $|\mathcal{C}|$                           & $\{100, 150, 200, 250, 5000\}$         &                    &                    & $250$              & $5000$             & $5000$             \\
Learning rate                                          & $\{5^{-3}, 10^{-3}, 5^{-4}, 10^{-4}\}$ & $5^{-4}$           & $5^{-4}$           & $5^{-4}$           & $5^{-4}$           & $5^{-4}$           \\
network (\#Layers $\times$ Units)                      & $\{2, 3\} \times \{128, 200, 500\}$    & $3 \times 200$     & $3 \times 200$     & $2 \times 128$     & $2 \times 128$     & $2 \times 128$ \\
Critic network (\#Layers $\times$ Units)               & $\{2, 3\} \times \{128, 200, 500\}$    & $3 \times 500$     & $3 \times 500$     & $2 \times 128$     & $2 \times 128$     & $2 \times 128$ \\
\bottomrule
\end{tabular}
\end{table}

\subsection{Recommender Systems}

\subsubsection{Baselines}
\begin{itemize}
    \item \textbf{SVD++} \citep{koren2008factorization}: An extension of the SVD algorithm popularised by Simon Funk during the Netflix Prize compeition, taking into account implicit ratings. For all experiments, we used the implementation provided in \cite{Surprise}.
    \item \textbf{Multitask MLP}: See previous Section.
    \item \textbf{MAML}: See previous Section.
\end{itemize}

\subsubsection{Experimental details}

We now discuss details of the experiment on recommender systems. For the 100k dataset, we use the provided movie genre (a k-hot vector), rating time stamps (normalised to mean zero and standard deviation 1) and user features (age, sex, occupation) as well as a learnable movie embeddings as context information. As we adapt to unseen users at test time, we do not learn a user-specific embedding or provide the user id as input to the model. At test time, we use the train sets estimates of the empirical mean and standard deviation of time stamps to normalise time stamps for test users. For the 20m dataset, we also provide a low-dimensional representation of a sparse tag matrix $\mathbf{T}$ where $T_ij$ is the occurrence of the j-\textit{th} tag to the i-\textit{th} movie. As this is an extremely large matrix, we follow \citep{strub2016hybrid} and perform PCA keeping the 50 greatest eigenvectors and normalising them by their square-root of the respective eigenvalue. Note that user features are not available for the 20m dataset. Note that as certain unpopular movies have only received a handful of ratings, we map all movies in the training set with only a single occurrence to a specific shared embedding for such low-resource ratings. At test time, this allows us to predict previously unobserved movies by using this shared embedding. Note that this is similar to how out-of-vocabulary words are handled in Natural Language processing. Choices for architecture and hyperparameters are provided in Tables \ref{table:movielens_hparams_100k} and \ref{table:movielens_hparams_20m}. Shown is both the range of hyperparemters considered for each model as well as the best values used to report results in the main text. For all models, we used Adam as the Optimiser during pre-training and SGD for gradient-based task inference.

Note that for both datasets, we split the dataset into 70\% training, 10\% validation and 20\% test users. After finding all hyperparameters on the validation set, we report test set results after training on validation and test set. Note that is non-standard: As opposed to withholding a fraction of ratings for \textit{known users}, we reserve all associated ratings for those users, testing the recommender system under more realistic conditions. To allow for reproducability of the results, we report all users ids for (randomly chosen) test set users here: \url{https://tinyurl.com/yyfzlg2x}.

For the Information Gain criterion:

\begin{equation}
\label{eq:info_gain}
    \mathcal{IG}(a_i) := \mathcal{H}\big(p(\mathbf{r}_{\setminus i}|\mathbf{a}_{\setminus i}, \mathcal{C})\big) - \mathbb{E}_{\hat{r}_i \sim p(\mathbf{r}_{i}|a_{i}, \mathcal{C})} \big[\mathcal{H}\big(p(\mathbf{r}_{\setminus i}|\mathbf{a}_{\setminus i}, \mathcal{C} \cup \{a_i, \hat{r}_i\})\big)\big]
\end{equation}

we arrive at the following arm selection strategy:

\begin{align}
    a^* &= \argmax_{a_i}{\mathcal{IG}(a_i)} \nonumber \\
        &= \argmin_{a_i} \mathbb{E}\big[\mathcal{H}\big(p(\mathbf{r}_{\setminus i}|\mathbf{a}_{\setminus i}, \mathcal{\mathcal{C}} \cup \{a_i, \hat{r}_i\})\big)\big] \nonumber \\
        &= \argmin_{a_i} \mathbb{E}\big[ \frac{1}{2}\ln(|\Sigma|) + \frac{|\mathcal{\mathcal{T}}|-1}{2}(1+\ln(2\pi))\big] \nonumber \\
        &= \argmin_{a_i} \mathbb{E}\big[\ln(\prod_j \sigma^2_j) \big] 
\end{align}

where we made use of conditional independence, the analytic form of the entropy of a multivariate normal and the determinant of a diagonal matrix. We thus seek to recommend the next item such that the product of variances of all other items in the target set given the user's expected response is minimised. For the results shown using this criterion, we used $5$ samples to estimate $\mathbb{E}_{\hat{r}_i \sim p(\mathbf{r}_{i}|a_{i}, \mathcal{C})}\big[\mathcal{H}\big(p(\mathbf{r}_{\setminus i}|\mathbf{a}_{\setminus i}, \mathcal{C} \cup \{a_i, \hat{r}_i\})\big)\big]$ for the 100k dataset and merely 1 sample for MovieLens-20m due to the much larger set of items.

\begin{table}[]
\footnotesize
\centering
\caption{Hyperparameters for the experiments in Movielens-100k. Attention \citep{kim2018attentive} indicates whether an attentive neural process was used. Decoder type indicates what type of MLP was used as the decoder. Options are standard MLPs, Skip connections \cite{dieng2018avoiding} and Residual networks \citep{he2016deep}. $\mathcal{C} \subset \mathcal{T}$ denotes whether context datapoints are included in the target set. We also introduce a minimum value for the standard deviation of predictions at target points. Hyperparameters for baselines models can be found in \citep{chen2018federated}.}
\label{table:movielens_hparams_100k}
\begin{tabular}{lll}
\hline
\textbf{Parameter}                & \textbf{Considered range} & \textbf{Neural Process}                            \\ \hline
Encoder (\#Layers $\times$ Units) & $\{2,3,4\}\times\{16, 32, 64\}$  & $3\times16$  \\
Decoder (\#Layers $\times$ Units) & $\{2,3,4\}\times\{16, 32, 64\}$  & $3\times16$  \\
Batch size                        & \{1, 16, 32\}                             & 32                               \\
Movie Embedding size              & \{8, 16, 32\}                             & 8                                \\
Meta Learning rate                & \{$10^{-4}$, $5\cdot10^{-5}$, $10^{-5}$\} & $10^{-5}$                                \\ \hline
Attention                         & \{None, Laplace, Multihead\}              & Multihead                                \\
Decoder type                      & \{None, Skip, ResNet\}                    & Skip                                      \\
$\mathcal{C} \subset \mathcal{T}$ & \{True\}                                  & True                                  \\
Min $\sigma$                      & \{0.1\}                                   & 0.1                                   \\ \hline
\end{tabular}
\end{table}

\begin{table}[]
\footnotesize
\centering
\caption{Hyperparameters for the experiments on Movielens-20m. Shown are both the range of considered values as well as the best option for each model. \textbf{NP-specific parameters}: Attention \citep{kim2018attentive} indicates whether an attentive neural process was used. Decoder type indicates what type of MLP was used as the decoder. Options are standard MLPs, Skip connections \citep{dieng2018avoiding} and Residual networks \citep{he2016deep}. $\mathcal{C} \subset \mathcal{T}$ denotes whether context datapoints are included in the target set. We also introduce a minimum value for the standard deviation of predictions at target points.}
\label{table:movielens_hparams_20m}
\begin{tabular}{@{}llllll@{}}
\toprule
\textbf{} & \textbf{Parameter}           & \textbf{Considered Range}             & \textbf{MAML}         & \textbf{Multitask MLP} & \textbf{Neural Process} \\ \midrule
          & Encoder (\#Layers $\times$ Units)     & $\{2, 3\}\times\{128\}$      & $2\times128$ & $2\times128$  & $2\times128$   \\
          & Decoder (\#Layers $\times$ Units)     & $\{2, 3\}\times\{128\}$      & $2\times128$ & $2\times128$  & $2\times128$   \\
          & Meta Learning rate                    & \{$10^{-4}, 10^{-5}\}$       & $10^{-4}$    & $10^{-4}$     & $10^{-5}$      \\
          & Update Learning rate                  & \{$10^{-2}, 10^{-3}\}$       &              &               &                \\
          & Inner gradient steps                  & \{1, 5, 10\}                 & 10           &               &                \\
          & Number of training tasks              & \{2000\}                     & 2000         & 2000          & 2000           \\
          & Rollouts per training task            & \{10\}                       & 10           & 10            & 10             \\
          & Rollout length                        & {50, 100, 200}               & 100          & 100           & 100            \\ \midrule
          & Attention                             & \{None, Laplace, Multihead\} &              &               & Laplace        \\
          & $\mathcal{C} \in \mathcal{T}$         & \{True\}                     &              &               & True           \\
          & min $\sigma$                          & \{0.1\}                      &              &               & 0.1            \\
          & Fixed sigma                           & \{False\}                    &              &               & False          \\
          & Maximum context size: $|\mathcal{C}|$ & $\{50, 100, 200, 300, 350\}$ &              &               & 300            \\ \bottomrule
\end{tabular}
\end{table}

For completeness we also include results for the more common dataset split, where ratings of all users are available during training. This allows for a comparison to more competitive Recommender System algorithms. We consider the same hyperparameters as reported in Table \ref{table:movielens_hparams_20m}, with the exception of attention which we found not to be useful and hence removed from the model. In addition, we included trainable user embeddings of size 128 and dropout with 20\% drop probability on both user and movie embeddings. Note that we observed signs of overfitting with this model which may be be resolved by more careful choice or regularisation. However, we would like to stress that the results reported in the main text are under more realistic conditions and thus of more interest.

\begin{table}[ht]
\small
\centering
\caption{Results on MovieLens 20m using an alternative dataset split more common in the literature. We report results on 10\% unseen ratings of \textit{known} users. In both cases, we report the RMSE. Baseline results taken from \citep{strub2016hybrid}.}
\begin{tabular}{lc}
\toprule
\textbf{Model}         & \textbf{90\%} \\
\midrule
BPMF \citep{salakhutdinov2008bayesian}              & 0.8123 \\
SVDFeature \citep{chen2012svdfeature}               & 0.7852 \\
LLORMA \citep{lee2013local}                                               & 0.7843 \\
ALS-WR \citep{zhou2008large}                                              & 0.7746 \\
I-Autorec \citep{sedhain2015autorec}                & 0.7742 \\
U-CFN \citep{strub2016hybrid}                                             & 0.7856 \\
I-CFN \citep{strub2016hybrid}                       & \textbf{0.7663} \\
\midrule
NP (random)                                                              & 0.7957 \\
\bottomrule
\end{tabular}
\label{table:movielens}
\end{table}

\subsection{Adversarial Task Search}
\label{appendix:adversarial}

\subsubsection{Baselines}
\begin{itemize}
    \item \textbf{BBB} (Bayes by Backprop) \citep{blundell2015weight}: Mean-field variational inference approximation to a posterior over neural network weights.
    \item \textbf{$\alpha$-Div} \cite{hernandez2016black}: An alternative Bayesian Neural network that provides a smooth interpolation between variational Bayes and an algorithm similar to expectation propagation (EP) by changing the parameter $\alpha$.
    \item \textbf{GP} (\citep{rasmussen2003gaussian}: A Gaussian Process with a linear and Matern 3/2 product kernel. No approximations are made. 
    \item \textbf{DKL} (\citep{wilson2016deep}): A Gaussian Process with the kernel applied to the final activations of a neural network. All weights are learned through standard optimisation of the kernel hyperparameters. We used the same linear and Matern 3/2 product kernel.
\end{itemize}

All baseline implementations with the exception of DKL are taken from \citep{riquelme2018deep}. Note that in order to provide a fair comparison, we account for the lack of pre-training by applying the above methods directly on embeddings of start/goal positions and the map layout, which leads to significant improvements in comparison to the application on raw data.

\subsubsection{Experimental details}

The task of interested is \texttt{explore\_goal\_locations\_large} from DMLab-30~\citep{beattie2016dmlab}. We consider a set of $K=1000$ randomly generated mazes $\{M_{1}, \ldots, M_{K}\}$, such that for each maze $M$ there is only a finite set (of capacity $C=1620$) of possible (around 40 each) agent and goal positions. This comes from the fact that in this task, start and goal positions can only appear in certain parts of the map (such as rooms) but not in corridors for instance.

We consider four types of agents: IMPALA~\citep{espeholt2018impala}, PopArt~\citep{hessel2018panultimate}, MERLIN~\citep{wayne2018merlin} and R2D2~\citep{kapturowski2018recurrent}. The hyperparameters of each agents are taken from the corresponding publications. For each agent type, we train four instances, each with standard Multitask-learning on four randomly sampled DMLab-30 levels in addition to \texttt{explore\_goal\_locations\_large}, the level of interest. Each agent is trained for approximately 100 hours until convergence, though the exact training time differs slightly. This difference due to the fact that the training was stopped for some of the agents because it converged.

Thereafter, for each map together with start and goal position, each agent is evaluated for $30$ episodes (in a complex 3-D environment, which takes about $10$ minutes in total on one local machines), and the final performance is reported as the mean over all 30 episode returns. Thus, we arrive at a dataset containing $1000 \times 1620$ examples for each of the $16$ agents. The data collection is done using a distributed using the MapReduce~\citep{dean2008mapreduce} framework.

We randomly split the mazes as well as the agents into 80\% training and 20\% holdout sets, which makes 800 training and 200 holdout mazes, and 12 training agents (3 of each type) and 4 holdout ones (1 of each type). During the pre-training of the Neural Process, the holdout mazes and agents are never observed.

In the main text we discussed that it is possible to decrease the complexity of the full search problem $\mathcal{O}(NKC)$ to $\mathcal{O}(\frac{N}{l} (K + lC))$ by proposing the promising maze and by reusing the solution of \textit{position search}. For this purpose, we use an auxiliary \textit{maze model} $g: M \rightarrow \mathbb{R}$, which for a given maze $M$ directly predicts the minimum reward over all possible agent and goal position. This model is a surrogate model which context set represents all the agent function evaluations on different mazes and positions, and it predicts the global minimum of the agent on given maze.

\begin{table}[]
\footnotesize
\centering
\caption{Hyperparameter selection for the Neural Process pretraining on the adversarial task problem. The values min/max/fixed $\sigma$ correspond to the standard deviation of the predictive distribution at target points. In case fixed $\sigma$ is None, we allow $\sigma$ to be learned within the interval $[\min \sigma, \max \sigma]$. We also consider \textit{Squared Exponential} (Sq.Exp.) attention, which replaces the \textit{Laplace} kernel with squared exponential.
}
\label{table:agi_pretrain_np_hparams}

\begin{tabular}{@{}llll@{}}
\toprule
\textbf{Parameter}            & \textbf{Considered Range}             & \textbf{Position Model} & \textbf{Maze model} \\ \midrule
Encoder (\#Layers $\times$ Units)                       & $\{3, 4\} \times \{128\}$             & $ 3 \times 128$     & $ 4 \times 128$ \\
Encoder (\#Layers $\times$ Units)                      & $\{3, 4, 5, 7\} \times \{128\}$       & $ 7 \times 128$     & $ 4 \times 128$ \\
Context size: $|\mathcal{C}|$ & $\{50, 100, 300, 400, 500, 600\}$     & $300$                   & $100$               \\
Latent dim. $\dim(z)$         & $\{16, 32, 64, 128, 256\}$            & $128$                   & $64$                \\
Attention                     & \{None, Sq.Exp., Laplace, Multihead\} & Sq.Exp                  & None                \\
Attention scale               & $\{0.01, 0.1, 0.5, 1.0\}$             & $0.5$                   &                     \\
Decoder Type                  & \{None, Skip\}                        & Skip                    & Skip                \\
$\mathcal{C} \in \mathcal{T}$ & \{True, False\}                       & False                   & False               \\
min $\sigma$                  & $\{0.1\}$                             & $0.1$                   & None                \\
max $\sigma$                  & $\{None, 20.0, 30.0, 60.0, 80.0\}$    & $30.0$                  & None                \\
fixed $\sigma$                & $\{None, 0.01, 0.1, 0.5, 1.0, 2.0\}$  & None                    & $0.1$ \\             
\bottomrule
\end{tabular}
\end{table}

We pre-train a Neural Process for both the \textit{position search} and \textit{full search} problems. For each hyperparameter, the Neural process is trained for $10^6$ iterations. For the \textit{position search} problem, each batch element corresponds to a fixed agent and map returns over 1620 positions. For the \textit{full search} problem, we consider up to 150 points in the batch element containing data from different maps and positions but for a fixed agent. The hyperparameter ranges and the best values are given in Table~\ref{table:agi_pretrain_np_hparams}.

For the Bayesian Optimisation experiments, we evaluate the model for each holdout agent, on all $200$ holdout mazes using $10$ random simulations (i.e. different initial context points, random seeds to control stochasticity etc) for the \textit{position search}, and $100$ random simulations for the \textit{full search}. The results are then averaged and we report the mean and the standard deviation of the scaled current minimum:

\begin{equation}
    \label{eq:cur_min}
    \hat{f}^{m}_{t}(M) = \frac{ f^{m}_{t}(M) - f^{m}(M)}{f^{M}(M) - f^{m}(M)},
\end{equation}
where $f^{m}_{t}(M)$ is a current minimum of the agent performance on the maze $M$ up to the iteration $t$, $f^{m}(M)$ and $f^{M}(M)$ are the global minimum and maximum of the agent on the maze $M$ (over all
positions).

\begin{table}[]

\footnotesize
\centering
\caption{Hyperparameter selection for baseline training on the \textit{Position search} problem.}
\label{table:agi_eval_hparams_pos}

\begin{tabular}{@{}llllll@{}}
\toprule
\textbf{Parameter}            & \textbf{Considered range}                            & \textbf{BBB}           & \textbf{$\alpha$-Div}  & \textbf{GP} & \textbf{DKL}           \\ \midrule
Learning rate                 & $\{10^{-1}, 10^{-2},5^{-3},10^{-3},5^{-4},10^{-4}\}$ & $10^{-2}$              & $10^{-2}$              & $10^{-3}$   & $5^{-3}$               \\
Training frequency            & $\{1, 2, 5, 10, 20\}$                                & $5$                    & $5$                    & $5$         & $5$                    \\
Training epochs               & $\{100, 200, 500, 1000, 2000\}$                      & $1000$                 & $1000$                 & $1000$      & $1000$                 \\
$\alpha$                      & $\{0, 0.1, 0.3, 0.5, 0.7, 0.9, 1.0\}$                &                        & $1.0$                  &             &                         \\
Decoder (\#Layers $\times$ Units)                      & $\{1, 2, 3\} \times \{100, 128, 256\}$               & $3 \times 100$         & $3 \times 100$         &             & $2 \times 128$ \\
Decoder-variance $\sigma$     & $\{0.1, 0.5, 1, 5, 10, 15, 20\}$                     & $1.0$                  & $20.0$                 &             &                        \\
Prior-variance $\sigma_{p}$   & $\{0.1, 0.5, 1, 5, 10, 15, 20\}$                     &                        & $20.0$                 &             &                         \\
Initial variance $\sigma_{0}$ & $\{0.1, 0.3, 1.0, 5.0, 10.0\}$                       & $0.3$                  & $0.3$                  &             &                        \\ \bottomrule
\end{tabular}
\end{table}

\begin{table}[]

\footnotesize
\centering
\caption{Hyperparameter selection for baseline training on the \textit{Full search} problem.}
\label{table:agi_eval_hparams_full}

\begin{tabular}{@{}llllll@{}}
\toprule
\textbf{Parameter}            & \textbf{Considered range}                            & \textbf{BBB}           & \textbf{$\alpha$-Div}  & \textbf{GP} & \textbf{DKL}             \\ \midrule
Learning rate                 & $\{10^{-1}, 10^{-2},5^{-3},10^{-3},5^{-4},10^{-4}\}$ & $10^{-2}$              & $10^{-2}$              & $10^{-1}$   & $10^{-3}$                \\
Training frequency            & $\{1, 2, 5, 10, 20\}$                                & $5$                    & $5$                    & $5$         & $5$                      \\
Training epochs               & $\{100, 200, 500, 1000, 2000\}$                      & $1000$                 & $1000$                 & $200$       & $200$                    \\
$\alpha$                      & $\{0, 0.1, 0.3, 0.5, 0.7, 0.9, 1.0\}$                &                        & $1.0$                  &             &                          \\
Decoder (\#Layers $\times$ Units)                      & $\{1, 2, 3\} \times \{64, 100, 128, 256\}$           & $3 \times 100$         & $3 \times 100$         &             & $ 3 \times 64$ \\
Decoder-variance $\sigma$     & $\{0.1, 0.5, 1, 5, 10, 15, 20\}$                     & $1.0$                  & $20.0$                 &             &                          \\
Prior-variance $\sigma_{p}$   & $\{0.1, 0.5, 1, 5, 10, 15, 20\}$                     &                        & $20.0$                 &             &                          \\
Initial variance $\sigma_{0}$ & $\{0.1, 0.3, 1.0, 5.0, 10.0\}$                       & $0.3$                  & $0.3$                  &             &                          \\ \bottomrule
\end{tabular}
\end{table}

For the \textit{full search} case, we do $l=5$ position iterations for Neural process and $l=2$ for the Random search.

\nocite{nagabandi2018deep}
\nocite{kim2018attentive}

\bibliography{submission}
\bibliographystyle{abbrvnat}

\end{document}